\pdfoutput=1
\documentclass[11pt, a4paper, logo, copyright]{googledeepmind}

\pdftrailerid{redacted}

\makeatletter
\renewcommand\bibentry[1]{\nocite{#1}{\frenchspacing\@nameuse{BR@r@#1\@extra@b@citeb}}}
\makeatother

\usepackage{kantlipsum, lipsum}
\usepackage{dsfont}
\usepackage{gdm-colors}
\usepackage[utf8]{inputenc}   
\usepackage{newunicodechar}   
\usepackage{amssymb}          
 
\usepackage[authoryear, sort&compress, round]{natbib}

\usepackage{bbding}
\usepackage[utf8]{inputenc} 
\usepackage[T1]{fontenc}    
\usepackage{hyperref}       
\usepackage{url}            
\usepackage{booktabs}       
\usepackage{nicefrac}       
\usepackage{microtype}      
\usepackage{amsmath}
\usepackage{graphicx}
\usepackage{fontawesome5}
\usepackage{multicol}
\usepackage[nameinlink]{cleveref}
\usepackage{bbm}
\usepackage{multirow}
\usepackage{soul}
\usepackage{float}
\usepackage{wrapfig}
\usepackage{blindtext}
\usepackage{tablefootnote}
\usepackage{amsfonts}
\usepackage[flushleft]{threeparttable}
\usepackage{colortbl}
\usepackage{mathtools,amssymb}
\usepackage{bm}
\usepackage{makecell}
\usepackage{caption}
\usepackage{capt-of}
\usepackage{array}
\usepackage{calc}      
\usepackage{caption}   
\usepackage{subcaption}  
\usepackage{xcolor,colortbl}
\usepackage[bottom]{footmisc}

\usepackage{xspace}

\newcommand{\eat}[1]{}

\crefformat{section}{\S#2#1#3}

\graphicspath{{Figures/}}

\title{KwaiMind Technical Report}

\renewcommand{\today}{}

\author{%
    \parbox{\linewidth}{\centering
        \fontsize{9}{12}\selectfont
        KwaiMind Team, Kuaishou Group\par\vspace{6pt}
        {\normalfont\fontsize{10}{12}\selectfont
        \href{https://github.com/KwaiMmu/KwaiMind}{%
            \textcolor{black}{\faGithub}\hspace{0.5em}%
            \textcolor{blue!60!black}{\texttt{https://github.com/KwaiMmu/KwaiMind}}}}
    }%
}

\begin{abstract}

Instruction-based image editing has made rapid progress, yet commercial content production demands more than general instruction following and visual quality. Generated images must preserve product identity, render promotional text accurately, and appeal to users. We present KwaiMind, an image editing system that combines broad editing competence with domain-specific capabilities for e-commerce. KwaiMind integrates an agent-based data engine, a staged adaptation pipeline, and a commercial benchmark. The data engine coordinates filtering, targeted generation, hierarchical annotation, and quality auditing to maintain approximately 1.8 million high-quality editing pairs. Using a multimodal diffusion transformer (MMDiT), we perform continued pre-training and supervised fine-tuning on general and e-commerce data, followed by preference optimization and online reinforcement learning. A general-purpose vision-language judge is complemented by specialized rewards for click-through rate (CTR), text rendering, and fine-grained product consistency. We optimize these objectives through specialized policies and consolidate their capabilities into a single editor via on-policy distillation. To assess practical utility, we introduce Ecom-Bench, covering 11 commercial editing tasks with task-specific visual evaluation and CTR-based ranking. KwaiMind achieves the strongest overall scores among the evaluated open-source editors on ImgEdit, GEdit, both language splits of REDEdit, and Ecom-Bench visual quality, while attaining the highest aggregate CTR ranking score among the compared systems. Offline, CTR-guided optimization increases the proportion of generated images whose predicted CTR exceeds that of the original product image from 12.16\% to 37.41\%. In an online A/B experiment, CTR-based selection of product main images yields an approximately 2.44\% relative increase in actual CTR. These results demonstrate the value of combining domain-specific data, reward-driven alignment, and commercially grounded evaluation for practical image editing.

\end{abstract}

\begin{document}
\maketitle

\begin{figure}[H]
    \centering
    \includegraphics[width=\textwidth,height=0.30\textheight,keepaspectratio]{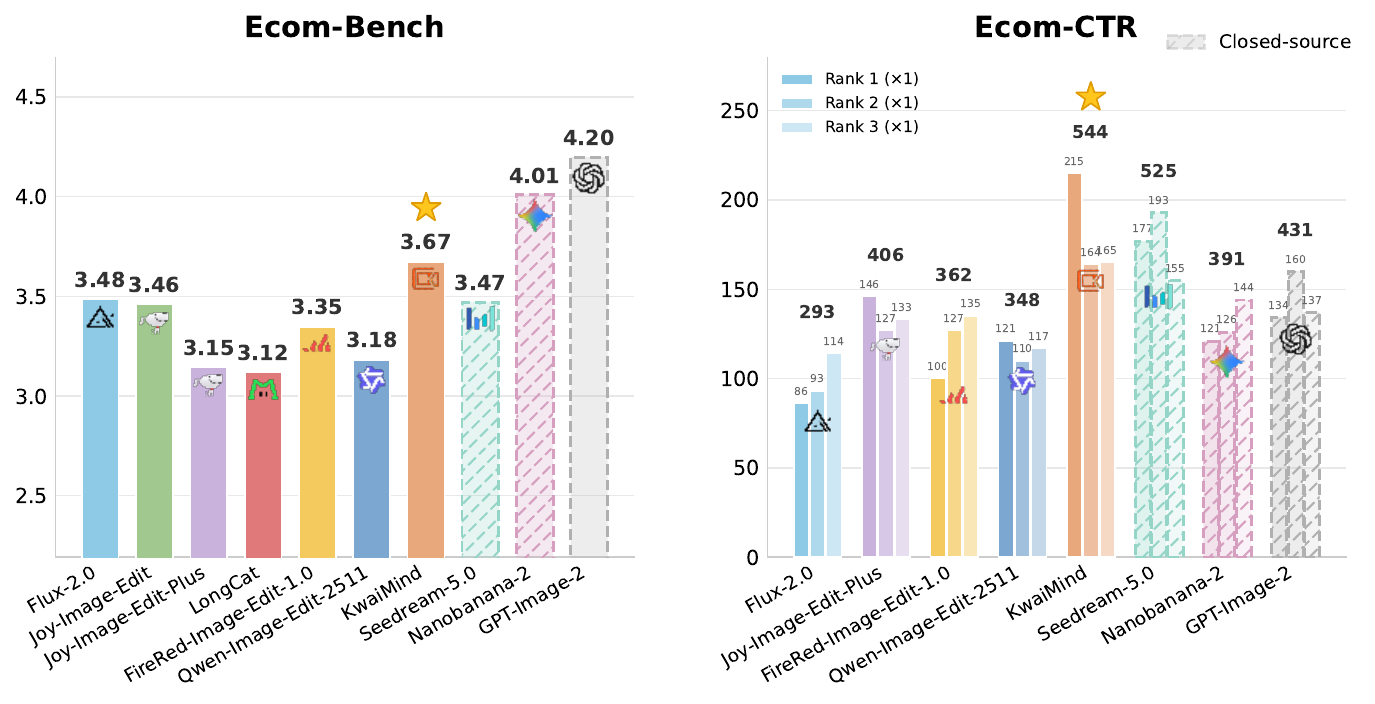}
    \caption{Overall comparison on Ecom-Bench visual quality and Ecom-CTR ranking. Hatched bars denote closed-source models.}
    \label{fig:bench-ecom}
\end{figure}

\newpage
\setcounter{tocdepth}{2} 

\tableofcontents 

\clearpage

\clearpage
\begin{figure}[H]
    \centering
    \includegraphics[width=0.98\textwidth]{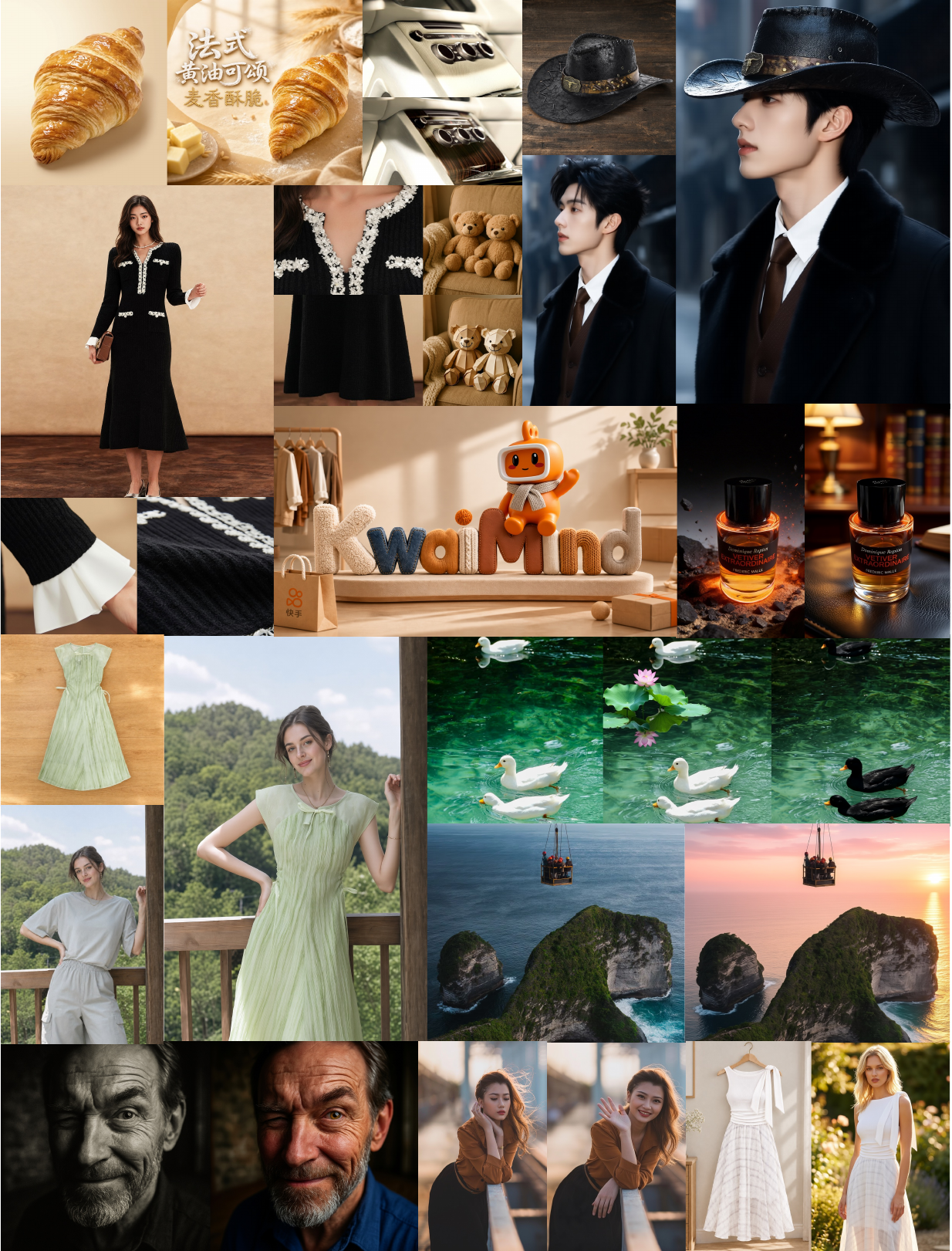}
    \caption{Showcases of KwaiMind across multiple image-editing tasks.}   
    \label{fig:showcases}
\end{figure}
\clearpage

\newpage

\section{Introduction}
\label{sec:introduction}

\begin{figure*}[t]
    \centering
    \includegraphics[width=\textwidth]{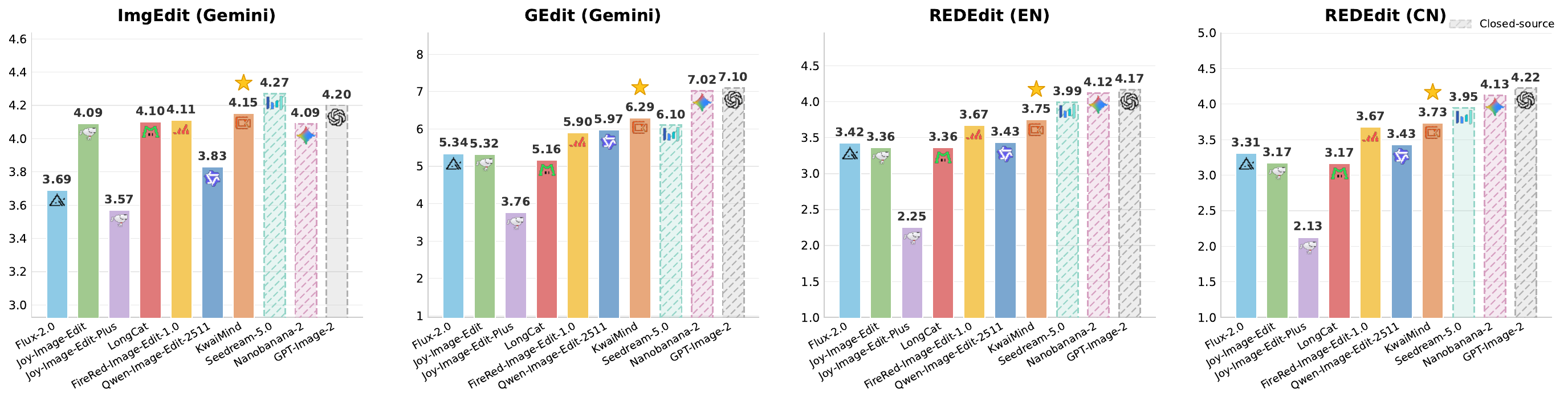}
    \caption{Overall comparison on general image editing benchmarks: ImgEdit, GEdit, and the English and Chinese splits of REDEdit. Hatched bars denote closed-source models.}
    \label{fig:bench-general}
\end{figure*}

Instruction-based image editing has advanced rapidly, and modern diffusion editors can now follow open-domain natural-language instructions to add, remove, replace, restyle, and recompose visual content with high fidelity~\citep{brooks2023instructpix2pix, labs2025flux, wu2025qwen, wei2025omniedit, yu2025anyedit}. Diffusion models also have broad applications in other domains~\citep{gong2026sculpting,li2025diffpcn,xia2025diffpc,xia2026diric,chang2026lvocal, lin2026joint}, while e-commerce image production offers a particularly valuable application of instruction-driven editing. On a modern e-commerce platform, every product may require many display images: a garment shown on different models and in different poses, a product placed against a variety of scenes, a promotional poster that highlights selling points, or a clean cutout for a catalog. Producing these images by hand is slow and expensive, and the volume grows with the size of the catalog, which makes automated, instruction-driven editing an attractive and valuable tool for merchants~\citep{yang2024cg4ctr,wang2021hybrid}.

Despite this promise, general-purpose editors do not directly meet the demands of commercial image production. E-commerce editing is governed by requirements that generic benchmarks neither isolate nor reward. The product itself must remain faithful across a large change of background, pose, or viewpoint, since any drift in shape, color, texture, or brand identity misrepresents the item being sold~\citep{li2026detailanywhere}. Text on product images, including prices, promotions, and selling points, must be rendered legibly and correctly, and even small glyph-level errors are immediately visible to shoppers~\citep{tuo2024anytext,wang2026textrefine}. Garment tasks such as virtual try-on impose their own constraints on fit, drape, and material fidelity, and they require reasoning jointly over more than one reference image~\citep{choi2024idmvton}. Above all, the ultimate measure of a commercial image is whether it attracts users, a property that is not fully captured by generic perceptual quality and that a general editor is not trained to optimize~\citep{wang2021hybrid,yang2024cg4ctr}. These requirements are largely absent from the data, the objectives, and the evaluation protocols on which general editing models are built.

We identify three obstacles that stand between a capable general editor and a production-grade e-commerce editing system. First, high-quality training data for the domain is scarce: e-commerce editing spans many specialized tasks, real product photography is noisy and unevenly distributed across categories, and constructing clean, instruction-aligned editing pairs at scale requires filtering, generation, and annotation far beyond what a fixed public corpus provides~\citep{wei2025omniedit,yu2025anyedit}. Second, standard training objectives do not target the domain-critical properties above. Supervised fine-tuning on imitation data teaches instruction following but does not directly optimize product identity preservation, text correctness, or commercial appeal. Third, existing benchmarks measure broad editing competence rather than the requirements of commercial image production, making it difficult to assess a model's suitability for real-world e-commerce workflows.

Addressing these challenges requires coordinated design across data construction, model alignment, and evaluation. We therefore introduce \textbf{KwaiMind}, an instruction-based image editing system for e-commerce that brings these elements together through three tightly connected components.

\paragraph{An agent-based data engine.}
We construct training data with a multi-agent pipeline that turns raw and generated imagery into clean, instruction-aligned editing pairs at scale (\cref{sec:data}). A Filter Agent enforces data quality through an online pre-filter over intrinsic image properties and a simulated-human post-filter that approximates human review, a Generation Agent repairs rejected samples and synthesizes data for under-covered tasks, a Caption Agent produces hierarchical instructions and verifies them through a reverse-audit loop, and a Coordinator Agent schedules the whole process as an auditable, closed-loop state machine with human intervention reserved for high-value and boundary cases. The engine produces domain-concentrated, quality-controlled data across the full range of e-commerce editing tasks.

\paragraph{A domain-adapted model with reward-driven alignment.}
Following the architectural design of Qwen-Image~\citep{wu2025qwen}, we build on a Multimodal Diffusion Transformer (MMDiT)~\citep{esser2024scaling} and adapt the model to the e-commerce domain through a staged pipeline of continued pre-training, supervised fine-tuning, and reinforcement learning from human feedback (\cref{sec:training}). The Data Agent maintains approximately 1.8M high-quality training pairs spanning general and e-commerce editing. Continued pre-training uses the subset above 720p, together with instruction augmentation, to develop high-resolution editing capabilities across both domains. Supervised fine-tuning then refines editing precision on a task-balanced corpus curated by the Data Agent, comprising 115.0K general editing pairs and 119.9K e-commerce pairs selected under stricter criteria for instruction accuracy, content preservation, and visual quality. The alignment stage is where the domain-critical objectives are optimized directly. We first apply Direct Preference Optimization~\citep{rafailov2023direct,wallace2023diffusionmodelalignmentusing} in a mixed offline and online regime, then run online reinforcement learning on the forward diffusion process, driven by a vision-language judge for general edits and by three dedicated reward models that target the properties general rewards miss: a click-through rate model for commercial appeal, a coarse-to-fine reward for visual text rendering, and a consistency reward for product and model identity preservation~\cite{li2026detailanywhere, wang2026cac}. Because each objective is optimized most effectively on its own, we finally consolidate the specialized policies into a single model through on-policy distillation, yielding one editor that inherits all capabilities without cross-task interference.

\paragraph{A commercial-grade benchmark.}
We introduce \textbf{Ecom-Bench}, an evaluation suite purpose-built for e-commerce image editing (\cref{sec:benchmark}). It covers 11 representative tasks spanning garment and wearable editing, composition and layout, text operations, and appearance transfer, and it scores each task with a behavior-anchored, per-task selection of general and e-commerce-specific dimensions under a geometric-mean protocol that penalizes any single critical defect. Beyond judge-based scores, Ecom-Bench reports a learned click-through rate score that estimates the commercial attractiveness of a generated image. Alongside Ecom-Bench, we evaluate on established general-domain image editing benchmarks.

The main contributions of this work are as follows.
\begin{itemize}
    \item We present an agent-based data engine that automates filtering, targeted generation, hierarchical captioning, and coordination into a closed-loop system for producing high-quality e-commerce editing data at scale.
    \item We adapt a strong open-source editor to the e-commerce domain with an alignment stage that combines mixed offline and online preference optimization, forward-process reinforcement learning, and three e-commerce reward models for commercial appeal, text rendering, and identity consistency, consolidated into a single model by on-policy distillation.
    \item We introduce Ecom-Bench, an 11-task benchmark with domain-tailored, behavior-anchored metrics and a learned click-through rate score, providing an evaluation protocol aligned with the requirements of commercial image production.
\end{itemize}

\section{Data}
\label{sec:data}

This section will describe the agent-based system for filtering, generating, annotating, and coordinating training data.

\subsection{Data Agent}
\label{sec:data-agent}

Data Agent is a multi-agent collaborative system for producing image-editing data. Coordinated by Coordinator Agent and jointly executed with the Filter, Generation, and Caption Agents, with human reviewers serving as a fallback at critical checkpoints, it transforms raw materials end-to-end into high-quality \textit{source-target-caption} triplets, while simultaneously producing the evaluator dataset used to iterate the simulated-human VLM evaluator.

\subsubsection{System Architecture}

The system comprises four collaborating sub-agents. All Skills are uniformly registered in the Skill Registry for dynamic invocation, and human reviewers are explicitly involved at critical checkpoints to handle boundary and failure cases.

\paragraph{Coordinator Agent.} The global scheduling hub. It maintains the sample lifecycle through a finite state machine, drives the sub-agents with standardized work orders, and manages loop budgets, gap-triggered supplementary generation, and the accumulation of strategy experience.

\paragraph{Filter Agent.} The data quality gatekeeper. Pre-Filter removes low-quality and non-compliant samples before annotation under two collaboration modes, Voter and Aspect. Post-Filter, applied after annotation, scores each sample along three criteria with a simulated-human VLM evaluator that iterates on itself.

\paragraph{Generation Agent.} The data remediation and completion component. The Modification branch repairs recoverable samples based on diagnostic reports, and the Supplement branch fills subtask gaps. An LLM Planner performs task parsing and Skill routing, and the actual generation is carried out by four categories of expert models.

\paragraph{Caption Agent.} The textual annotation component. It produces three-tiered hierarchical captions for each sample, uses difference-mask guidance to direct the VLM toward the edited regions, and verifies annotation quality through a closed-loop reverse audit performed by an LLM.

\subsection{Filter Agent}
\label{sec:filter-agent}

Filter Agent serves as the core component responsible for data filtering and quality control. It is divided into two stages according to its position in the pipeline. Pre-Filter focuses on intrinsic image quality, removing low-quality, non-compliant, or duplicate samples before images enter the annotation stage. Post-Filter simulates human review by performing fine-grained evaluation on data tuples with generated captions, approximating human review capability and continuously accumulating training data for downstream model iteration.

\subsubsection{Pre-Filter}
\label{sec:pre-filter}


\paragraph{Multi-Model Collaboration.}
Each Pre-Filter Skill uses either Voter or Aspect collaboration. Voter mode handles holistic judgments such as AIGC and anime detection or content compliance. It invokes $K$ heterogeneous VLMs in parallel and adopts the majority verdict; when no majority exists, a Judge model aggregates their verdicts and rationales \cite{wang2022self,zheng2023judging}. Aspect mode handles composite criteria such as perceptual quality, aesthetics, and product-task consistency. It decomposes each criterion into $N$ approximately orthogonal aspects evaluated by independent prompts or dedicated detectors, followed by weighted aggregation with a Judge model \cite{song2024finesure}. All Skills output an evidence chain and one of four verdicts: \textit{Passed}, \textit{Rejected-Recoverable}, \textit{Rejected-Unrecoverable}, or \textit{Uncertain}. These verdicts map to the \texttt{PreFilter-\{Passed, Recoverable, Rejected, Uncertain\}} states. At the Pre-Filter stage, \textit{Uncertain} denotes voter disagreement, contradictory aspect evidence, or low confidence.

\paragraph{Pre-Filter Skills.}
We implement four Pre-Filter Skills. Deduplication, basic visual feature filtering, consistency filtering, and the perceptual quality and aesthetics sub-Skill use Aspect mode. The AIGC and anime detection and content compliance sub-Skills use Voter mode.

\begin{enumerate}

    \item Deduplication. Global deduplication uses CLIP embeddings \cite{radford2021learning} for nearest-neighbor retrieval and clustering. Edit-pair deduplication combines CLIP similarity with PSNR, SSIM, and LPIPS \cite{zhang2018unreasonable} to remove near-identity pairs \cite{team2026firered}.

    \item Basic visual feature filtering. Thresholds on saturation, brightness, and RGB entropy remove under-exposed, over-exposed, color-distorted, and near-uniform images \cite{team2026firered}. For e-commerce data, CTR signals further exclude images with poor historical performance.

    \item Compliance filtering. This Skill comprises three sub-Skills. (1) AIGC and anime detection excludes images with a pronounced AI-generated appearance and anime-style images. (2) Perceptual quality and aesthetics assessment evaluates blur, noise, exposure, physical anomalies, composition, color harmony, watermarks, and text overlays. (3) Content compliance review uses multi-model voting to detect nudity, graphic violence, and other prohibited content and produces an auditable rejection report.

    \item Consistency filtering. For editing pairs involving products, persons, and related subjects, this Skill checks category, subject identity, and core visual attributes between source and target images.

\end{enumerate}

\paragraph{Pre-Filter Output and Routing.}
Each verdict includes routing metadata. \textit{Passed} records confidence; \textit{Rejected-Recoverable} records rejection reasons, recovery hints, and evidence chains; \textit{Rejected-Unrecoverable} records rejection categories; and \textit{Uncertain} records the disagreement source and aggregate confidence. Coordinator routes samples using these fields. Batch-level reports record pass rates, Skill-level rejection distributions, aspect scores, collaboration disagreement, and sampled positive and negative cases for data monitoring and Skill updates.

\subsubsection{Post-Filter}
\label{sec:post-filter}

\paragraph{Evaluation Dimensions.}
Post-Filter applies a simulated-human VLM evaluator to captioned tuples of a source image, target image, and caption. Following FireRed \cite{team2026firered}, it evaluates three dimensions. (1) Instruction consistency measures whether the source-to-target change correctly and completely follows the caption without extraneous edits. (2) Edit consistency measures preservation outside the intended edit region, including subject identity. (3) Perceptual quality measures sharpness, artifacts, and aesthetics. For each dimension, the evaluator outputs a score on a 1--10 scale, together with confidence and a rationale. It then assigns \textit{Accepted}, \textit{Rejected}, \textit{Uncertain}, or \textit{OOD}, corresponding to the \texttt{PostFilter-\{Accepted, Rejected, Uncertain, OOD\}} states. \textit{Uncertain} denotes an in-distribution sample with contradictory dimension scores or confidence below a preset threshold. \textit{OOD} denotes an input outside the evaluator's supported distribution. Both states are routed to human review.

\paragraph{Evaluator Training Data.}
Evaluator training data comes from four sources. Positive samples are high-confidence \textit{Accepted} outputs from the current evaluator. Auto-constructed negatives are generated by an LLM through controlled perturbations of attributes, subjects, operations, or categories in Accepted samples \cite{team2026firered}. Human samples include reviewed contrastive pairs and manually corrupted samples. Pre-Filter negatives are high-confidence, low-disagreement samples rejected by Pre-Filter in Section \ref{sec:pre-filter}. The evaluator is updated through prompt revision with high-value errors and accumulated experience as few-shot examples, and through supervised fine-tuning. Updates are manually deployed when the accumulated samples reach a preset threshold and offline validation shows no performance degradation.

\subsection{Generation Agent}
\label{sec:generation-agent}

Generation Agent performs Modification and Supplement. Modification repairs samples marked as \textit{Rejected-Recoverable} by Pre-Filter, while Supplement fills subtask gaps detected by Coordinator Agent. In both branches, an LLM Planner routes each work order to an expert model or API. The expert produces multiple candidates, from which the highest-scoring result is retained. Each output records its execution path, expert models, intermediate boxes or masks, and prompts. Modified samples enter the \texttt{Generated-Modified} state with a region mask for local Pre-Filter, whereas supplementary samples enter the \texttt{Generated-Supplement} state and undergo full-image Pre-Filter. Modification is limited to $N$ rounds, and samples exceeding this budget are routed to the human review queue.

\subsubsection{LLM Planner}

The LLM Planner converts each work order into an executable plan through three steps: (1) Diagnostic parsing. For Modification, it extracts repair targets from the rejection reasons, recovery hints, and evidence chain, including missing subjects, attribute mismatches, and unintended edits. For Supplement, it determines the target subtask, scene constraints, and diversity requirements from the gap list and dataset distribution. This step is omitted when the work order already specifies the target scene. (2) Skill routing. The Planner inserts localization or segmentation before spatially constrained tasks such as replace, remove, and change-background. It then selects a Skill according to task type, scene complexity, and the Skill capability profile. Expert models remain fixed within each Skill and are not exposed to the Planner. (3) Prompt generation. The Planner generates semantically equivalent and lexically diverse prompts for the selected Skill to reduce mode collapse \cite{brooks2023instructpix2pix}.

\subsubsection{Expert Execution}

For composite edits, the Skill layer follows the Task Splitting mechanism in FireRed \cite{team2026firered}. A VLM decomposes each instruction into ordered atomic operations, and the Router sequentially invokes the corresponding experts to reduce per-step complexity and preserve structure. We group these experts by control signal: (1) Instruction-driven editing. FLUX.2 \cite{flux-2-2025} and Qwen-Image-Edit-2511 \cite{wu2025qwen} handle general edits, LongCat-Image \cite{longcat2025image} handles dense Chinese text, and Seedream and NanoBanana-2 handle complex multimodal instructions. (2) Perception. GroundingDINO \cite{liu2024grounding}, SAM2 \cite{ravi2025sam}, and RMBG-2.0 \cite{rmbg2} provide bounding boxes, masks, and foreground separation for spatially constrained edits. (3) Structured-control editing. Mask-conditioned FLUX.2 and SDXL-Inpainting \cite{podell2024sdxl} perform localized edits, while DWPose \cite{yang2023effective} provides keypoints for pose transfer. (4) Deterministic synthesis. We use 3D parametric templates, structured layout templates, and deterministic image-processing operators \cite{team2026firered} for color transfer, sharpening, layout control, and parametric pose or expression control. This group also serves as a fallback and a supplementary data source.

\subsection{Caption Agent}
\label{sec:caption-agent}

Caption Agent annotates images and editing pairs marked as \textit{Passed} by Pre-Filter through hierarchical captioning, mask-guided prompting, and bounded reverse auditing. The VLM produces three captions for each sample: (1) Detailed caption. It describes the image content for text-to-image data or the attribute, spatial, and semantic changes in an editing pair. (2) Concise caption. An LLM compresses the detailed caption using randomly sampled syntactic structures from multiple VLMs and lexicons to reduce template bias \cite{singla2024pixels}. (3) Simulated-user caption. The LLM rewrites the concise caption as a colloquial, help-seeking instruction. For editing pairs, we compute an approximate change mask from the pixel-level difference between the source and target images. The mask is overlaid on the target image and encoded as Set-of-Mark boxes and indices in the prompt \cite{yang2023set}. This directs the VLM to describe edited regions, including small object replacements, local tagline removal, and text modification.

After captioning, a reverse audit checks semantic accuracy following a generate-then-verify procedure \cite{wu2026generate}. Given only the caption, an LLM infers the expected visual content and compares its key slots with the known metadata. Any mismatch is added to the prompt for caption regeneration. The loop terminates when the audit passes or reaches $N$ rounds \cite{madaan2023self}; remaining failures are forwarded by the Coordinator for human review.

\subsection{Coordinator Agent}
\label{sec:coordinator-agent}

Coordinator Agent maintains the global dataset state, schedules sample batches, and dispatches standardized work orders across sub-agents. It also manages loop budgets and human-review entry points. Each work order records its branch triggers for reproducibility and auditing.

\subsubsection{State Machine and Work Orders}

Coordinator tracks each sample through the finite state machine in Table \ref{tab:coordinator-states}. Every state transition emits a work order containing the source and target states, rejection reason, evidence chain, execution path, and metadata. The pipeline contains two feedback loops: Pre-Filter with Modification and Caption with Reverse-Audit. Both loops are capped at $N$ rounds. Samples exceeding the loop budget enter the corresponding stage-specific Escalated state and are routed to human review.

\begin{table}[ht]
\centering
\caption{Sample states maintained by Coordinator Agent. Each state determines the next work-order destination. States ending in \texttt{Escalated} indicate that the corresponding loop budget has been exhausted and human review is required.}
\label{tab:coordinator-states}

\begin{tabular}{ll}
\toprule
\textbf{State} & \textbf{Description} \\
\midrule
Ingested & Newly ingested sample \\
PreFilter-\{Passed, Recoverable, Rejected, Uncertain\} & Pre-Filter verdict \\
Generated-\{Modified, Supplement\} & Generation output \\
Captioned & Captioning completed \\
Modification-Escalated & Modification budget exhausted \\
Caption-Escalated & Reverse-audit budget exhausted \\
PostFilter-\{Accepted, Rejected, Uncertain, OOD\} & Post-Filter verdict \\
Evaluator-\{Human, Negative, Positive\} & Evaluator training data\\
Human-\{Approved, Rejected, Recoverable, Relabeled\} & Human verdict \\
Ingested-To-Trainset & Pending database entry \\
\bottomrule
\end{tabular}
\end{table}

\subsubsection{Scheduling Memory}

Coordinator maintains loop memory and an experience buffer. Loop memory records each sample's iteration count and rejection sequence for budget control and stopping. The experience buffer stores representative successes and failures, human verdicts, and automatically constructed negatives for Skill prompt updates and SFT training. This design combines per-sample loop state with cross-batch experience \cite{zhang2025survey}. For Supplement, Coordinator periodically checks subtask coverage, category distribution, and diversity against preset thresholds. Detected gaps trigger Supplement work orders to Generation Agent. All scheduling decisions follow a work-order-based ReAct loop \cite{yao2022react}.

\subsubsection{Human-in-the-Loop}

Human intervention occurs at three entry points: (1) Milestone review. After each high-cost stage, Coordinator reports the batch pass rate, Skill-level rejection distribution, subtask-level pass rate, and disagreement by collaboration mode. Representative accepted, rejected, and boundary samples are included for approval before the next stage. (2) Exception handling. \textit{Uncertain} samples from either filtering stage, Post-Filter \textit{OOD} samples, loop-budget overflows, and operational threshold violations are routed to human review. Reviewers resolve these cases and annotate boundary samples. (3) Iteration decisions. Reviewers select samples for the experience buffer, Skill prompt updates, or SFT training. They also trigger retraining and deployment of the simulated-human VLM evaluator.

\subsection{End-to-End Data Pipeline}
\label{sec:data-pipeline}

\begin{figure}[htbp]
    \centering
    \includegraphics[width=\linewidth]{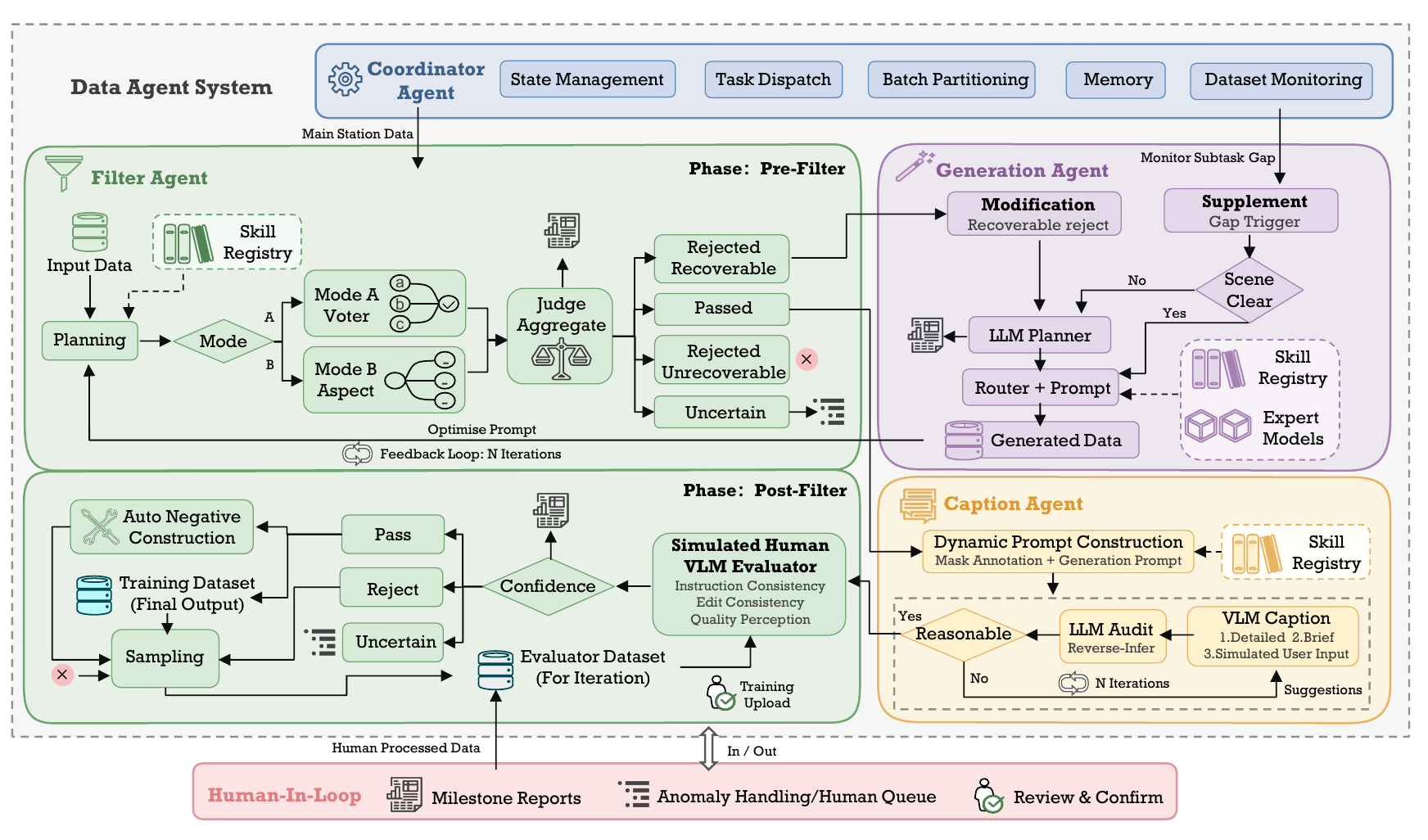}
    \caption{Overview of the end-to-end data pipeline. Coordinator Agent routes samples through filtering, generation, captioning, and post-filtering with bounded feedback loops and human review.}
    \label{fig:pipeline}
\end{figure}

Figure \ref{fig:pipeline} summarizes the end-to-end pipeline. Coordinator uses its state machine to route sample batches through Pre-Filter, Generation, Caption, and Post-Filter, with the Skill Registry providing shared capabilities. The pipeline produces the main training set and an evaluator dataset for iterative VLM evaluator updates.

Pre-Filter routes \textit{Passed} samples to Caption Agent, \textit{Rejected-Recoverable} samples with diagnostic reports to Modification, and \textit{Uncertain} samples to human review. \textit{Rejected-Unrecoverable} samples are archived, with high-confidence cases added to the evaluator dataset. Supplement addresses subtask gaps detected by Coordinator, and all Generation outputs return to Pre-Filter. Caption Agent applies a bounded generation, reverse-audit, and revision loop before qualified source-image, target-image, and caption triplets enter Post-Filter. High-confidence \textit{Accepted} samples enter the training-set ingestion queue, while high-confidence \textit{Rejected} samples and automatically perturbed hard negatives are archived as evaluator negatives. Post-Filter \textit{Uncertain} and \textit{OOD} samples are routed to human review. Accepted and rejected outputs also update the evaluator dataset. Coordinator enforces all loop budgets and routes milestone reports, exceptional or low-confidence samples, and on-demand inspections to human review. Human verdicts update pipeline states and evaluator data.

\subsection{Dataset Scale and Final Composition}
\label{sec:data-scale}

\begin{figure*}[t]
    \centering
    \includegraphics[width=\textwidth]{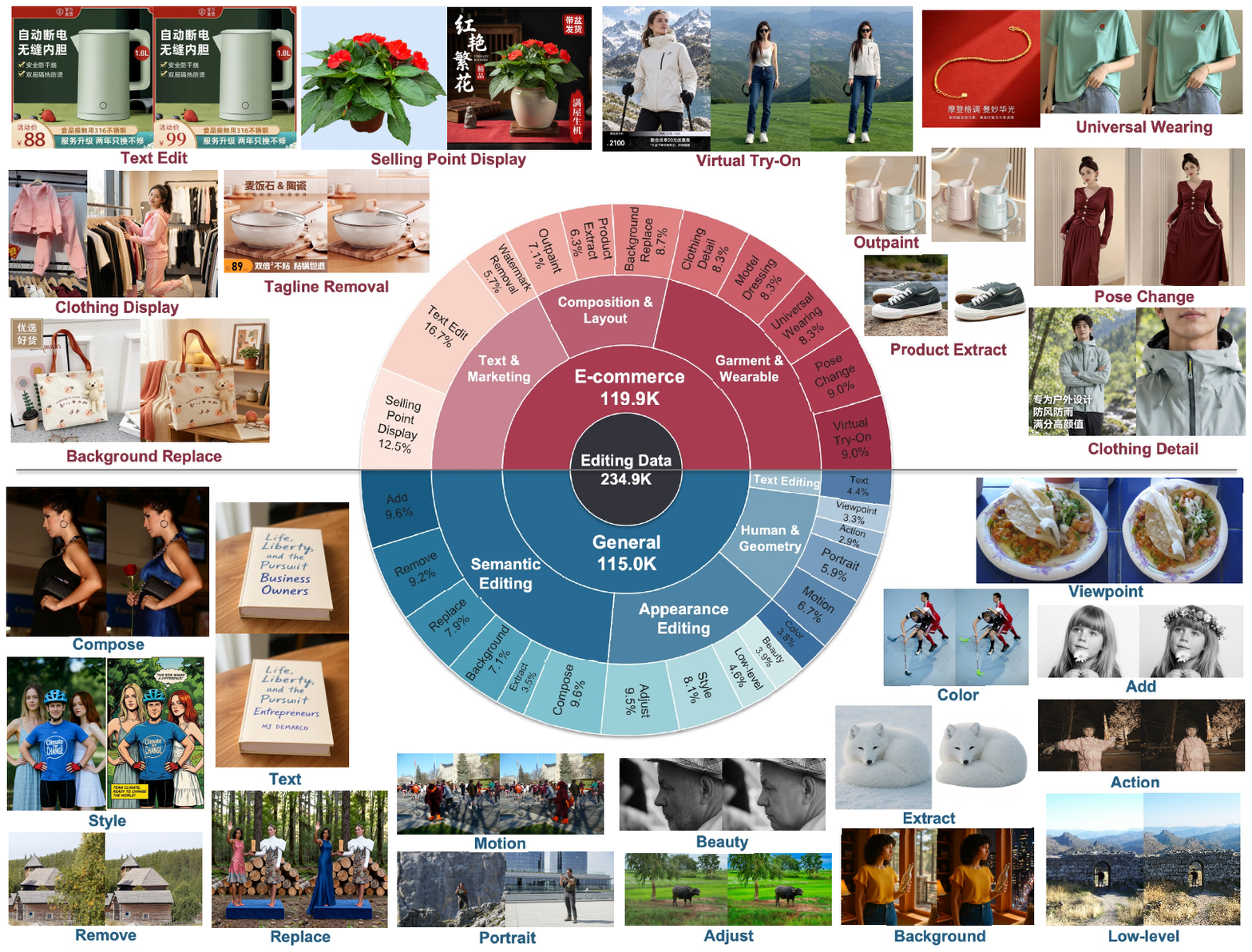}
    \caption{Composition of the curated SFT corpus. The upper half shows the
    e-commerce subset and the lower half shows the general editing subset; surrounding
    examples illustrate representative source--target pairs from different tasks.}
    \label{fig:data-distribution}
\end{figure*}

We apply the Data Agent to a mixture of public editing datasets and proprietary Kwai pairs. These sources contain approximately 26.2M source--target examples before the final quality-control and deduplication stages. Through filtering, targeted modification of recoverable samples, and gap-driven supplementary generation, the Data Agent maintains approximately 1.8M high-quality training pairs for model development.

\begin{table}[ht]
\centering
\caption{Data sources processed by the Data Agent and the resulting maintained training corpus.}
\label{tab:data-sources}
\small
\begin{tabular}{lr}
\toprule
\textbf{Data source} & \textbf{Number of samples} \\
\midrule
ScaleEdit-12M~\citep{chen2026scaleedit} & 12.0M editing pairs \\
X2Edit~\citep{ma2026x2edit} & 3.7M editing pairs \\
AnyEdit~\citep{yu2025anyedit} & 2.5M editing pairs \\
KwaiData & 8.0M editing pairs \\
\midrule
Pair-based source total & 26.2M pairs \\
Maintained training corpus & $\sim$1.8M pairs \\
\bottomrule
\end{tabular}
\end{table}

Candidate editing pairs that pass Pre-Filter and complete captioning and reverse auditing enter Post-Filter, where instruction consistency, edit consistency, and perceptual quality are each scored on a 1--10 scale as defined in \cref{sec:post-filter}. Following the routing in \cref{sec:data-pipeline}, high-confidence \textit{Accepted} triplets enter the training-set ingestion queue, while high-confidence \textit{Rejected} samples are archived as evaluator negatives. The maintained training pairs are then selected according to their downstream role: broad high-resolution pairs support CT, while more selective and task-balanced subsets support SFT and final alignment. Modification repairs samples identified as recoverable by Pre-Filter; Supplement addresses task or visual-pattern gaps detected by Coordinator. Outputs from both branches return to Pre-Filter and proceed through captioning and Post-Filter before they can enter the training corpus.

To prepare the data for supervised fine-tuning, we organize candidates from the maintained 1.8M-pair corpus by general and e-commerce editing tasks. We first filter candidate pairs using their recorded Post-Filter scores, then apply an additional round of task-specific screening with criteria tailored to each subtask across instruction accuracy, content preservation, and visual quality, alongside task balancing and quality auditing. This process yields 234.9K SFT pairs, comprising 115.0K general editing pairs covering atomic and compositional edits and 119.9K e-commerce pairs drawn from proprietary data collected from real merchant demands. The e-commerce subset spans 11 tasks grouped into three families: \textit{Garment \& Wearable}, \textit{Composition \& Layout}, and \textit{Text \& Marketing}. Figure~\ref{fig:data-distribution} shows the task distribution of the curated SFT corpus, and the detailed selection criteria are provided in \cref{sec:sft}.

Finally, we construct dedicated preference and reward-training subsets from the maintained corpus. We stratify e-commerce and general editing examples across the target tasks, while using additional expert-generated pairs when required to improve long-tail coverage. This staged reuse of the same quality-controlled pool supports CT, SFT, and final alignment without treating earlier filtered data as discarded.

\section{Training}
\label{sec:training}

We build our editing model on top of a strong open-source foundation~\cite{wu2025qwen} and adapt it through a staged pipeline comprising continued pre-training (CT), supervised fine-tuning (SFT), and reinforcement learning from human feedback~\citep{wu2026step}. CT combines instruction augmentation with a broad mixture of general and e-commerce editing pairs above 720p, prepared by the Data Agent, to develop high-resolution editing capabilities across both domains. SFT then refines instruction accuracy, content preservation, and visual quality using a task-balanced subset curated through score-based filtering and an additional round of screening with task-specific criteria across these three dimensions. The final alignment stage combines preference optimization, online reinforcement learning driven by task-specific reward models, and multi-task consolidation.

\subsection{Architecture}
\label{sec:training-architecture}

Our architecture follows Qwen-Image-Edit-2511~\citep{wu2025qwen} and is based on a double-stream Multimodal Diffusion Transformer (MMDiT)~\citep{esser2024scaling}. Three input streams are concatenated into a single token sequence for dense bidirectional attention: latent tokens of the target image produced by a variational autoencoder (VAE) encoder~\citep{rombach2022latent}, latent tokens of one or more reference images, and textual instruction embeddings produced by a vision-language encoder. Jointly processing the reference and target in a single stream enables each target token to attend to the reference content at every layer, a capability essential for preserving identity and layout in the editing tasks that predominate in commercial image production.

Position information follows a unified rotary scheme. Reference and target image tokens share a common spatial coordinate grid and are distinguished by a temporal offset, so that spatially corresponding regions of the reference and the edited result are placed in register while remaining separable by the model. Clean reference latents and noised target latents receive distinct time conditioning, which prevents the model from confusing the fixed reference with the signal it must denoise.

Because e-commerce editing frequently requires more than one reference, for example a model image together with a garment image in virtual try-on, the input stream accepts a variable number of reference images. Multiple references are encoded independently by the VAE encoder and appended to the sequence, each carrying its own positional and temporal tags. This paired multi-reference format is used consistently across training and inference so that the model learns to bind attributes from the correct source image.

\subsection{Continued Pre-Training} 
\label{sec:ct}

Continued pre-training (CT) adapts the foundation model using a broad mixture of general and e-commerce editing pairs from the approximately 1.8M maintained training pairs described in \cref{sec:data-scale}. These data are obtained through the Data Agent described in \cref{sec:data-agent}. From this automatically curated pool, we retain only image pairs with resolutions above 720p for CT, adapting the model to high-resolution editing scenarios where fine-grained textures, product details, and scene structure need to remain clear and consistent. Training on this high-resolution subset combines instruction augmentation with exposure to editing operations across both domains. E-commerce data is included from the outset, allowing the model to learn from diverse e-commerce products, materials, presentation scenes, and operation types alongside general editing tasks. This stage establishes broad, high-resolution editing competence across domains, providing the foundation for the more selective SFT stage.

\paragraph{Instruction augmentation.}
Building on the three instruction forms produced by the Caption Agent in \cref{sec:caption-agent}, we use Qwen3-VL-32B~\citep{bai2025qwen3vl} to provide Chinese and English versions of each form for every source--target pair. Detailed instructions specify the editing changes and constraints, concise instructions retain the essential editing intent, and simulated-user instructions express that intent as colloquial user requests. This yields a $2\times3$ instruction set: two languages combined with three expression styles. All six variants preserve the same editing intent and share the same visual supervision. During training, we sample one language and one instruction form for each pair in each epoch, encouraging robustness to differences in language, specificity, and phrasing rather than dependence on a single annotation style.

\subsection{Supervised Fine-Tuning} 
\label{sec:sft}

Supervised fine-tuning (SFT) refines the CT checkpoint on the training data curated by the Data Agent in \cref{sec:data-scale}. The 234.9K pairs comprise 115.0K general editing examples and 119.9K e-commerce examples, with their task distribution illustrated in Figure~\ref{fig:data-distribution}. The general subset covers atomic and compositional editing, while the e-commerce subset includes proprietary business data collected from real merchant demands. Whereas CT emphasizes broad exposure to domains and editing operations, SFT uses this selectively curated, task-balanced corpus to improve editing precision and output quality.

\paragraph{Task-wise selection criteria.}
Following the initial filtering based on Post-Filter scores, we apply an additional task-specific filter across three dimensions: instruction accuracy, content preservation, and visual quality. The screening criteria for each dimension are tailored to the editing requirements of individual subtasks, while sharing the following general principles:
\begin{itemize}
    \item \textbf{Instruction accuracy.} All requested modifications are completed, with no unintended changes beyond the instruction.
    \item \textbf{Content preservation.} Product design, color, logos, person identity, and non-edited regions remain faithful to the reference, except where a change is explicitly requested.
    \item \textbf{Visual quality.} Textures are clear, geometry is plausible, lighting and shadows are natural, and the output contains no conspicuous artifacts.
\end{itemize}

\paragraph{Task coverage and conditioning.}
The e-commerce subset spans 11 tasks grouped into \textit{Garment \& Wearable}, \textit{Composition \& Layout}, and \textit{Text \& Marketing}. We maintain coverage across these tasks and the general editing categories so that high-volume tasks do not dominate the curated mixture. For tasks requiring multiple references, we retain the multi-reference input format used at inference, enabling the model to associate each reference with the appropriate subject and preserve the relevant product and identity attributes.

\subsection{Reinforcement Learning with Human Feedback}
\label{sec:rlhf}

Supervised fine-tuning teaches the model to follow editing instructions, but it optimizes a maximum-likelihood objective on curated pairs and does not directly reward the properties that determine production quality, such as instruction faithfulness, identity preservation, text legibility, and commercial appeal. We therefore add a reinforcement learning from human feedback (RLHF) stage. Following the view that denoising can be treated as a multi-step decision process amenable to policy optimization~\citep{black2024ddpo}, we align the model against explicit reward signals rather than against a fixed reference distribution alone. Related applications of task-specific reinforcement learning span robotic control~\citep{wu2025arc}, multimodal reasoning and self-evolution~\citep{jiang2026vlm,heng2026eve}, and multi-view scene editing~\citep{wang2026geometry}.

\begin{figure}[ht]
    \centering
    \includegraphics[
        width=\textwidth,
        trim={0 0 0 0},
        clip
    ]{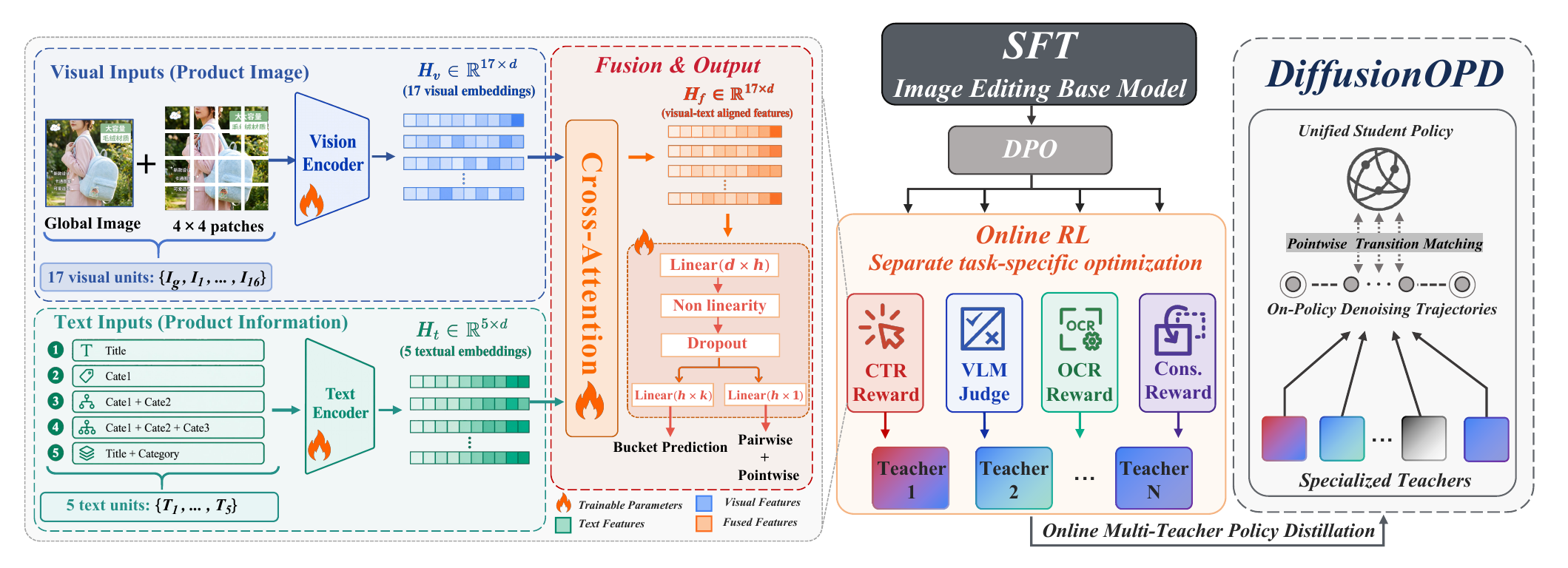}
    \caption{Overview of the proposed post-training framework.}
    \label{fig:rl_arch}
\end{figure}

Our alignment pipeline has three parts. We first apply Direct Preference Optimization (DPO), combining an offline phase on curated preference pairs with an online phase that regenerates pairs from the policy, to establish a stable preference-aligned checkpoint. We then run online reinforcement learning with DiffusionNFT~\citep{zheng2026diffusionnftonlinediffusionreinforcement}, whose reward is supplied by VLMs acting as task-conditioned judges and, for the e-commerce objectives that a general judge cannot score reliably, by three dedicated reward models targeting click-through rate, visual text rendering, and detail consistency. Because each of these objectives is optimized most effectively by its own reinforcement learning run, we finally consolidate the separately trained policies into a single model with DiffusionOPD~\citep{li2026diffusionopdunifiedperspectiveonpolicy}. The remainder of this subsection describes each component.

\subsubsection{DPO}
\label{sec:dpo}

The first alignment step is Direct Preference Optimization~\citep{rafailov2023direct} applied to the diffusion model~\citep{wallace2023diffusionmodelalignmentusing}. We train DPO in a mixed regime that combines an offline phase on a fixed preference corpus with an online phase that draws preference pairs from the model's own generations. The offline phase establishes a stable preference-aligned checkpoint from curated data, and the online phase keeps improving the policy against the distribution it actually produces, which mitigates the distribution shift that arises when a static offline set no longer matches the evolving model.

\paragraph{Offline DPO.}
The offline phase trains on a fixed corpus of preference pairs. For each editing condition $c$, which consists of the source image and the instruction, we draw several candidate edits from the supervised checkpoint and, where available, from expert editing pipelines. Each candidate is annotated for instruction compliance and visual quality, by human reviewers for a high-precision subset and by a vision-language judge for the remainder. Within the candidates for a given condition we take a high-scoring edit as the preferred sample $x^{w}$ and a low-scoring edit as the rejected sample $x^{l}$, so that both members of a pair share the same source and instruction and differ in editing quality. This same-condition construction isolates the quality signal from content differences and yields the fixed corpus $\mathcal{D}_{\mathrm{off}}$. We optimize the standard Diffusion-DPO \cite{wallace2023diffusionmodelalignmentusing} objective, which contrasts the current policy $\theta$ against a frozen reference policy $\theta_{\mathrm{ref}}$ initialized from the supervised checkpoint:
\begin{equation}
\label{eq:dpo}
\mathcal{L}_{\mathrm{DPO}}(\mathcal{D})
= - \mathbb{E}_{(c,\, x^{w},\, x^{l}) \sim \mathcal{D}}
\left[
\log \sigma \Big(
-\beta \big(
\Delta_{\theta,\mathrm{ref}}(x^{w}, c) - \Delta_{\theta,\mathrm{ref}}(x^{l}, c)
\big)
\Big)
\right],
\end{equation}
where $\sigma(\cdot)$ is the logistic function, $\beta$ is the preference temperature, and $\Delta_{\theta,\mathrm{ref}}(x, c) = \mathcal{L}_{\theta}(x, c) - \mathcal{L}_{\theta_{\mathrm{ref}}}(x, c)$ is the difference between the denoising losses of the current and reference policies on sample $x$. Minimizing \cref{eq:dpo} on $\mathcal{D}_{\mathrm{off}}$ raises the relative likelihood of preferred samples over rejected ones while the reference term regularizes the update toward the supervised checkpoint.

\paragraph{Online DPO.}
Offline DPO is bounded by its fixed corpus. As the policy improves during training, the stored pairs increasingly reflect mistakes the model no longer makes, the preferred edits fall below what the current policy can already produce, and the gradient from \cref{eq:dpo} weakens. To keep the preference signal aligned with the model's present behavior, we continue with an online phase that regenerates pairs from the policy itself. Treating denoising as a multi-step decision process and training on the model's own samples has been shown to optimize downstream rewards more effectively than reweighting a fixed dataset~\citep{black2024ddpo}, which motivates moving from a static corpus to on-policy pairs. At each round the current sampling policy $\pi_{\mathrm{old}}$ produces a group of candidate edits for a sampled condition $c$; each candidate is scored by the task-conditioned reward introduced in \cref{sec:diffusionnft}, that is a vision-language judge for general edits and the dedicated reward model of \cref{sec:ctr-reward,sec:ocr-reward,sec:consistency-reward} for the corresponding e-commerce objective. We form an on-policy pair by taking the highest-scoring candidate as $x^{w}$ and the lowest-scoring candidate as $x^{l}$, giving a continually refreshed set $\mathcal{D}_{\mathrm{on}}$ that targets the current failure modes. The sampling policy $\pi_{\mathrm{old}}$ is refreshed periodically from $\theta$ so that generation tracks the improving model, while the DPO reference $\theta_{\mathrm{ref}}$ remains the supervised checkpoint so that the regularizer keeps anchoring the policy and prevents the online updates from drifting too far from a trusted initialization~\citep{wallace2023diffusionmodelalignmentusing}.

\paragraph{Mixed training.}
We optimize the offline and online objectives jointly, so that the curated corpus keeps supplying reliable, human-grounded preferences while the on-policy pairs continually adapt to the evolving policy:
\begin{equation}
\label{eq:dpo-mix}
\mathcal{L}_{\mathrm{DPO}\text{-}\mathrm{mix}}
= \mathcal{L}_{\mathrm{DPO}}(\mathcal{D}_{\mathrm{off}})
+ \lambda_{\mathrm{on}}\, \mathcal{L}_{\mathrm{DPO}}(\mathcal{D}_{\mathrm{on}}),
\end{equation}
where $\lambda_{\mathrm{on}}$ balances the online term against the offline term. We warm up on the offline corpus alone and then anneal $\lambda_{\mathrm{on}}$ from $0$ to $0.5$ as on-policy pairs accumulate,
so that the online signal takes over only once the policy is reliable enough to generate informative pairs.

Mixed DPO gives a stable, preference-aligned checkpoint, but it still reduces each group of candidates to a single best-versus-worst pair and supervises the model through a binary comparison, discarding both the graded magnitude of the reward and the information in the remaining candidates. We therefore hand off to online reinforcement learning with DiffusionNFT, which consumes the full graded reward over every candidate in the group and optimizes it directly on the forward diffusion process. In effect, DPO provides a robust warm start from paired preferences, and DiffusionNFT extracts a finer, continuous learning signal from the same on-policy generations once the policy is strong enough to benefit from it.

\subsubsection{DiffusionNFT}
\label{sec:diffusionnft}

We adopt DiffusionNFT~\citep{zheng2026diffusionnftonlinediffusionreinforcement} for online reinforcement learning from the DPO checkpoint. It optimizes the forward diffusion process through flow matching~\citep{lipman2023flow}, using clean generated images and their rewards without estimating likelihoods or retaining denoising trajectories. For each editing condition $c$, the sampling policy generates a group of candidates whose rewards are normalized into optimality probabilities $r \in [0,1]$. The optimization objective is
\begin{equation}
\label{eq:nft}
\mathcal{L}_{\mathrm{NFT}}
= \mathbb{E}_{t,\, x_0 \sim \pi_{\mathrm{old}},\, \epsilon}
\left[
r \, \big\| v_{\theta}^{+}(x_t, c, t) - v \big\|_2^2
+ (1 - r) \, \big\| v_{\theta}^{-}(x_t, c, t) - v \big\|_2^2
\right],
\end{equation}
where $x_t$ is obtained by adding noise $\epsilon$ to a generated sample $x_0$, and $v$ is the target velocity. The implicit policies satisfy $v_{\theta}^{\pm} = v_{\mathrm{old}} \pm \beta\,[v_{\theta} - v_{\mathrm{old}}]$, with all velocity fields evaluated at the same input and $\beta$ controlling guidance strength. The sampling policy $v_{\mathrm{old}}$ is frozen during each update and refreshed by an exponential moving average of $v_{\theta}$. We use the VLM judge in \cref{sec:vlm-judge} for general editing quality and combine it with the corresponding CTR, OCR, or consistency reward for each specialized objective.

\subsubsection{VLM as Judge}
\label{sec:vlm-judge}

We use Gemini 3.1 pro preview directly through its API as a fixed vision-language judge without additional fine-tuning. The judge receives the source image, any additional reference images, the editing instruction, and the generated result. A common rubric defines three integer scores from 1 to 5 for image quality, instruction alignment, and aesthetics. These dimensions assess rendering integrity, editing correctness, and visual presentation, respectively. The judge evaluates each dimension independently and provides a brief justification grounded in visible evidence.

\paragraph{Image quality.}
Rendering fidelity is evaluated locally within the edited region and globally across the image. The assessment considers clarity and structural integrity, including geometric distortions, texture discontinuities, and inconsistencies in lighting or occlusion. Scores reflect the severity and spatial extent of these defects, with higher values indicating cleaner rendering and more coherent integration of the edit. Judgments are made within the intended visual style.

\paragraph{Instruction alignment.}
To assess instruction alignment, the judge compares the generated result with the source and reference images under the specified editing instruction. The comparison checks the requested operations, target attributes, and spatial relations, together with the preservation of subject identity and content outside the intended edit. Preservation is judged relative to the intended transformation, allowing changes necessary to carry out the instruction. Missing requirements, incorrect targets, and unrelated alterations reduce the score, which jointly reflects editing correctness and completeness.

\paragraph{Aesthetics.}
The aesthetic assessment considers how composition, color, and tone jointly organize the visual presentation. Framing, spacing, and contrast are examined to determine whether they establish a clear focal subject and a balanced relationship with the background. In product images, this assessment places particular emphasis on product visibility and the arrangement of supporting elements. High scores require these elements to form a cohesive presentation, with harmonious color and tonal relationships, consistent styling, and minimal distraction from the intended subject.

\subsubsection{CTR Reward Model}
\label{sec:ctr-reward}

General editing quality does not capture whether a product image will attract clicks in a live storefront~\citep{wang2021hybrid,yang2024cg4ctr}. Commercial appeal is a platform-specific signal that a general vision-language judge cannot score reliably, so we train a dedicated click-through rate (CTR) reward model that predicts the relative appeal of a candidate product image.

\paragraph{Data Collection and Filtering.}
We build the training set from platform behavior logs. For each product we gather the set of display images that have been served to users together with their logged impressions and clicks, and we form the empirical CTR of an image as its click count divided by its impression count. Two properties of raw logs make this signal noisy. An image with few impressions has a high-variance CTR estimate, and images of different products are not directly comparable because CTR is confounded by product category, price, and demand. We therefore filter the logs in two ways. First, we retain only images with more than 2,000 impressions to reduce noise in CTR estimates caused by low impression counts. Second, we build training examples only from pairs of images of the same product, which cancels product-level confounders and reduces the problem to ranking presentations of one item. We keep a pair only when its CTR ordering is consistent across all three observation windows of 7, 14, and 30 days, filtering out pairs whose ordering changes across these windows due to short-term traffic fluctuations.

\paragraph{Model Construction.}
The reward model encodes visual and textual product information at multiple granularities. On the visual side it encodes the full image together with a grid of local patches, which lets the model attend both to global composition and to local regions such as the product foreground and any rendered text. On the textual side it encodes the product title together with its multi-level category path. The visual and textual streams are encoded independently and fused by cross-attention, and a lightweight head maps the fused representation to a scalar score $s$. Encoding the two modalities separately before fusion avoids diluting the visual signal, which we find to be the primary carrier of commercial appeal.

\paragraph{Training Schemes.}
We train the reward model in two stages on the filtered same-product pairs. In the first stage we discretize observed CTR into $K$ buckets and train a bucket classifier with cross-entropy, which gives the encoder a coarse but stable notion of click attractiveness before it is exposed to fine-grained ranking. In the second stage we refine the model with a gap-weighted pairwise ranking loss combined with a pointwise regression term:
\begin{equation}
\label{eq:ctr-loss}
\mathcal{L}_{\mathrm{CTR}}
= \frac{1}{|\mathcal{P}|} \sum_{(i,j) \in \mathcal{P}} w_{ij}\, \log\!\big(1 + e^{-(s_i - s_j)}\big)
+ \lambda \, \frac{1}{N} \sum_{i=1}^{N} \big(s_i - \hat{c}_i\big)^2,
\qquad
w_{ij} = \min\!\left(1, \frac{|c_i - c_j|}{\gamma}\right),
\end{equation}
where $s_i$ is the predicted score, $c_i$ the observed CTR, $\hat{c}_i$ the regression target, $\mathcal{P}$ the set of same-product pairs with $c_i > c_j$, $\gamma$ a gap normalization constant, and $\lambda$ the regression weight. The pairwise term follows the logistic ranking formulation of \citet{burges2005learning} and teaches the model the ordering within each product, and the pointwise regression term keeps the absolute scores calibrated across products. The gap-aware weight $w_{ij}$ down-weights near-tie pairs, whose ordering is dominated by measurement noise, and emphasizes pairs with a clear CTR difference, which acts as an implicit curriculum. We set $K = 10$ buckets and $\lambda = 0.2$.

\paragraph{Method Comparison.}
We first compare the learned reward model against using a general-purpose vision-language model as a zero-shot CTR judge. General judges perform close to chance on pairwise CTR ranking, because commercial appeal is a platform-specific signal that is not recoverable from generic visual-text pretraining, whereas the dedicated model attains roughly $70\%$ pairwise ranking accuracy on our held-out pairs.
To position the model against prior work under a common protocol, we further evaluate on the public CreativeRanking benchmark~\citep{wang2021hybrid}, a large-scale creative-ranking dataset of over $1.7$M ad creatives from $500$K products. Because this benchmark provides only image input, we reduce our model to a vision-only variant. Even in this reduced form the model reaches $57.1$ pairwise accuracy, surpassing both general-purpose baselines (Qwen3-VL $47.9$, GPT-4o $50.2$, LLaVA $52.8$) and task-specific creative-ranking methods (VAM~\citep{wang2021hybrid} $52.2$, CG4CTR~\citep{yang2024cg4ctr} $53.1$, CAIG~\citep{chen2025caig} $56.2$), as summarized in \cref{tab:ctr-creativeranking}. This confirms that the training strategy, which first learns CTR buckets and then refines pairwise rankings, is the primary source of the model's ranking ability, rather than multimodal fusion alone.

\begin{table}[t]
\centering
\caption{Pairwise ranking accuracy on the public CreativeRanking benchmark. Our
vision-only variant (vision encoder plus MLP head, with the text branch and fusion
module removed) is compared against general-purpose and task-specific baselines.}
\label{tab:ctr-creativeranking}
\begin{tabular}{lc}
\toprule
\textbf{Model} & \textbf{PairAcc} \\
\midrule
Qwen3-VL            & 47.9 \\
GPT-4o              & 50.2 \\
Qwen3-VL-Trained    & 51.2 \\
VAM                 & 52.2 \\
LLaVA               & 52.8 \\
CG4CTR              & 53.1 \\
CAIG                & 56.2 \\
\midrule
Ours (Vision-only)  & \textbf{57.1} \\
\bottomrule
\end{tabular}
\end{table}

The model also holds a clear efficiency advantage. It scores an image with a single forward pass through a compact SigLIP2-base encoder~\citep{tschannen2025siglip2} and a lightweight head, which is orders of magnitude cheaper than prompting a multi-billion-parameter vision-language judge once per candidate, and this cost difference is decisive when the model is used both as an online training reward over large candidate groups and as an offline filter across the full production catalog. Scaling the encoder further brings little benefit: replacing SigLIP2-base with SigLIP2-giant improves pairwise accuracy by only about $0.2$ points, which indicates that the accuracy comes from the architecture and the training strategy rather than from encoder capacity, so a small, fast encoder is sufficient in practice. Ablations further show that the visual encoder is the primary information bottleneck, that overly fine bucket discretization introduces label noise, and that the soft gap-aware weight outperforms hard filtering of near-tie pairs. Taken together, these results justify a purpose-built, compact reward model over a prompted general judge for the commercial objective.

\paragraph{Effect on Generated Image Usability.}
We use the CTR reward model to score and select product images. In an online A/B experiment, this selection yields an approximately $2.44\%$ relative increase in observed CTR over the control group. We evaluate how reinforcement learning with the CTR reward changes the usability of generated product images on a fixed test set. For each source image and editing instruction, the editing models before and after CTR reward optimization each produce a candidate under the same generation protocol. The original product image serves as the common control, and each model's output is separately paired with this control. We define CTR-based usability as the fraction of evaluated pairs in which the generated image receives a strictly higher CTR than its original counterpart. Ties do not count as improvements, and both checkpoints are evaluated over the same test cases. Under this criterion, the usable proportion increases from $12.16\%$ before CTR reward optimization to $37.41\%$ afterward, an absolute gain of $25.25$ percentage points. Both rates measure improvement over the original product images, rather than direct wins between the two editing models. This result shows that CTR reward optimization increases the fraction of generated candidates preferred to the original images. This is an offline measure of generated image usability. Together, the offline generation evaluation and the online selection experiment support the CTR model's utility in guiding image generation and identifying visual assets with greater click potential.

\subsubsection{OCR Reward Model}
\label{sec:ocr-reward}

Product images frequently carry dense text such as prices, promotions, and selling points, and errors in rendered text are immediately visible to shoppers~\citep{tuo2024anytext}. String-level recognition rewards are insensitive to glyph-level defects such as missing strokes or distorted characters that harm legibility without changing the recognized string. We therefore adopt a coarse-to-fine text reward that combines a span-level term for semantic placement with a glyph-level term for structural fidelity~\citep{wang2026textrefine}.

The span-level reward evaluates whether the intended text spans appear at scene-appropriate locations without spurious or missing content. Detected text is first filtered to remove detections that overlap existing source text or the foreground product region, and each target span is matched to a detection through a normalized edit distance similarity. The span reward is the product of a fidelity term over matched spans and a coverage term that penalizes both missing target spans and spurious detections:
\begin{equation}
\label{eq:ocr-span}
R_{\mathrm{span}}
=
\frac{1}{N}\sum_{i=1}^{N} r_i
\cdot
\max\!\left(
0,\,
1-\frac{U_{\mathrm{tar}}+U_{\mathrm{ocr}}}{\max(N,|V|)}
\right).
\end{equation}

where $r_i$ is the per-span similarity of the $i$-th of $N$ target spans, $U_{\mathrm{tar}}$ and $U_{\mathrm{ocr}}$ are the numbers of unmatched target and detected spans, and $V$ is the set of valid detections. To mitigate reward hacking under prolonged optimization and preserve the appearance of the rest of the image, we add a gated structural regularizer that activates only once span accuracy is sufficiently high:
\begin{equation}
\label{eq:ocr-text}
R_{\mathrm{text}}
= R_{\mathrm{span}} + \mathbbm{1}\!\left[R_{\mathrm{span}} > \tau_{\mathrm{ssim}}\right] \cdot \mathrm{SSIM}\big(\psi(\hat{x}, V),\, \psi(x_{\mathrm{src}}, V)\big),
\end{equation}
where $\psi(\cdot, V)$ fills the retained text boxes in $V$ with white before the comparison, so that the SSIM term~\citep{wang2004ssim} measures structural preservation of the non-text regions alone rather than of the edited text itself, and $\tau_{\mathrm{ssim}}$ gates the regularizer so that span correctness remains the primary objective.

The glyph-level reward provides dense supervision on character structure. For a target glyph with an annotated character box, the region is cropped and passed to an OCR recognizer, and the reward is the maximum posterior assigned to the target glyph across the recognizer's output timesteps:
\begin{equation}
\label{eq:ocr-glyph}
R_{\mathrm{glyph}} = \max_{1 \le t \le T} P_{t,\, \mathrm{id}(y^{\star})},
\end{equation}
where $P \in [0,1]^{T \times |C|}$ is the recognizer's per-timestep character posterior matrix and $\mathrm{id}(y^{\star})$ is the vocabulary index of the target glyph. This graded signal decreases smoothly under stroke omission or structural distortion, which a binary recognition reward cannot express. The two rewards are applied by task type on a mixed stream, with the span-level reward supervising text insertion and the glyph-level reward supervising text replacement, and both are converted to the optimality probability consumed by \cref{eq:nft}.

\subsubsection{Fine-grained Consistency Reward Model}
\label{sec:consistency-reward}

Many e-commerce edits must preserve the identity of a product or a model across a substantial change of background, pose, or viewpoint, and pixel reconstruction losses do not capture failures such as color drift, implausible textures, or subtle identity deviation. A reward that only checks whether the overall subject looks like the same product is also insufficient, because the defects that matter commercially are often local: a mismatched cuff, a wrong button, a fabric weave that does not correspond to the source, or a detail region that is rendered plausibly but is not the part that was requested~\citep{li2026detailanywhere}. We therefore design a consistency reward that operates at two levels of granularity, and this multi-level formulation is a central contribution of our reward design. Concretely the reward scores a reference and a generated image along two complementary axes and combines them into a single scalar:
\begin{equation}
\label{eq:consistency}
r = \alpha_{\mathrm{id}}\, r_{\mathrm{id}} + \alpha_{\mathrm{tf}}\, r_{\mathrm{tf}},
\end{equation}
where the two axes capture, respectively:
\begin{itemize}
    \item \textbf{Product identity consistency} ($r_{\mathrm{id}}$): whether the result depicts the exact same product as the reference, verified against unique, non-generic features such as the specific fabric material, stitching and seam structure, and hardware or trims, rather than mere category or color similarity.
    \item \textbf{Target-part and detail fidelity} ($r_{\mathrm{tf}}$): whether the intended local part is depicted accurately and completely, with its local structure, spatial relationships, and fine process and decorative details preserved from the reference.
\end{itemize}
The reward thus verifies consistency not only at the level of the overall product subject, but also at the level of the requested part and of the fine-grained details within it. This part- and detail-level supervision is what distinguishes our reward from generic identity-oriented preference models, and it directly targets the local mismatches that determine whether a commercial detail image is usable.

\paragraph{Reward model and training data.}
The reward model is a vision-language model fine-tuned to emit the two integer scores under a fixed, behavior-anchored rubric, taking the reference image, the candidate image, and the prompt as input. Each of the two axes is defined by an explicit 1-to-4 rubric with positive and negative indicators, and the rubric enforces a strict-uncertainty principle: generic similarity alone cannot earn a high score, and any uncertainty about feature matching or part accuracy must be resolved toward the lower score. This makes the learned reward conservative, which is desirable for an RL signal that would otherwise be easy to hack with superficially plausible but inconsistent edits.

We build the training corpus by re-annotating a large pool of reference-candidate pairs, covering both high-quality detail shots and deliberately imperfect generations, so that the model sees the full range of consistency levels rather than only near-perfect examples. A stronger vision-language model scores every pair on the two axes under the same rubric, and we retain only the integer scores as supervision. Two filtering steps keep the labels reliable: pairs on which the annotator's rationale contradicts its score are discarded, and a held-out subset is checked so that the model separates known-consistent from known-inconsistent pairs on the identity and part-fidelity axes before it is used as a reward. We fine-tune the model for a small number of epochs on this filtered corpus.

During reinforcement learning the composite reward $r$ is normalized within each candidate group into the optimality probability consumed by \cref{eq:nft}, and the reference term retained from the earlier alignment stages keeps the update from eroding the model's editing ability.

\subsubsection{DiffusionOPD}
\label{sec:diffusionopd}

The capability-specific objectives described above are optimized through separate reinforcement learning runs. Each objective is defined by a composite reward that combines the general-purpose VLM judge with one specialized reward model. These objectives pair the VLM judge with the corresponding specialized reward models. The VLM judge provides a shared assessment of general output quality and reduces the risk of overoptimizing exploitable patterns in any single specialized reward. Separate optimization is preferable because jointly training on heterogeneous objectives can lead to cross-task interference and imbalance in optimization difficulty. Sequentially optimizing these objectives with a single policy can instead cause catastrophic forgetting. We therefore obtain one specialized teacher for each composite task objective and consolidate the teachers into a unified model using DiffusionOPD \citep{li2026diffusionopdunifiedperspectiveonpolicy}, an on-policy distillation method for multi-task diffusion training.

DiffusionOPD separates task-specific exploration from multi-task integration. In the first stage, we train task-specific teachers using DiffusionNFT. Each teacher is optimized for one composite objective and jointly considers the VLM judge and the corresponding specialized reward model. This design preserves the specialization induced by the task-specific reward while maintaining a shared notion of general output quality across all teachers. In the second stage, we distill the teachers into a single student along trajectories generated by the student itself.

Let $k$ index the task-specific teachers. Since the student and each teacher use the same denoising schedule, their one-step transition kernels are Gaussian distributions with a shared covariance and differ only in their means. The per-step reverse Kullback--Leibler divergence therefore reduces to the following mean-matching objective.
\begin{equation}
\label{eq:opd}
\mathcal{L}_{\mathrm{OPD}}
=
\mathbb{E}_{k \sim p(k)}
\mathbb{E}_{x_{0:N} \sim p_{S,\theta}^{k}}
\left[
\sum_{j=0}^{N-1}
\frac{
\left\|
\mu_{S}^{k}(x_{t_j};\theta)
-
\mu_{T_k}(x_{t_j})
\right\|_2^2
}{
2\bar{\sigma}_j^2
}
\right],
\end{equation}
where $\mu_{S}^{k}$ denotes the student transition mean for task $k$, $\mu_{T_k}$ denotes the transition mean of the corresponding teacher, and $\bar{\sigma}_j^2$ is the shared variance at denoising step $j$. The trajectories are sampled on-policy from the student under the data distribution of each task. Each teacher therefore supervises the states that the student actually visits for the corresponding task. During each training round, we collect the distillation loss for every task and apply one update using the aggregated loss. This balanced update prevents the consolidated student from being dominated by any single task objective.

\section{Benchmark}
\label{sec:benchmark}

To address the requirements of both practical business scenarios~\citep{wang2026textrefine,li2026detailanywhere,fan2026autopp} and general-purpose image editing~\citep{wei2025omniedit,yu2025anyedit,xia2025dreamomni}, we need a reliable evaluation protocol that characterizes model behavior across domain-specific commercial applications and broad editing tasks. We therefore evaluate KwaiMind in two domains: (1)~the \textbf{general domain}, using established image editing benchmarks that measure broad instruction-following ability, and (2)~the \textbf{e-commerce domain}, using our 11-task Ecom-Bench with domain-tailored evaluation criteria.

\subsection{General Image Editing Benchmark}
\label{sec:general-bench}

We assess general editing competence on three complementary public benchmarks: ImgEdit-Bench~\cite{imgedit}, which evaluates instruction following and visual quality with task-adaptive scores; GEdit-Bench~\cite{geditbench}, which combines semantic consistency and perceptual quality using VIEScore~\cite{viescore}; and REDEdit-Bench~\cite{team2026firered}, which assesses diverse editing tasks with parallel Chinese and English instructions, evaluated separately. ImgEdit averages dimension scores per sample and then across editing categories, while GEdit uses the geometric mean of semantic consistency and perceptual quality. Because the original ImgEdit and GEdit judges produced unstable scores across repeated evaluations, we use Gemini 3.1 Pro Preview for both, keeping their evaluation prompts and scoring rules unchanged.

\subsection{Ecom-Bench: E-commerce Image Editing Benchmark}
\label{sec:ecom-bench}

General-purpose editing benchmarks do not reflect the distinctive requirements of commercial image production. Tasks such as virtual try-on, selling-point poster composition, garment texture replication, and tagline removal must jointly meet requirements for product identity fidelity, typographic accuracy, and commercial visual appeal. General benchmarks neither isolate nor measure these criteria. We therefore introduce \textbf{Ecom-Bench}, a benchmark purpose-built for the e-commerce domain, covering 11 editing tasks under a systematic and domain-tailored evaluation protocol.

\subsubsection{Task Taxonomy}
\label{sec:ecom-tasks}

Ecom-Bench organizes e-commerce image editing into three thematic groups spanning 11 tasks in total:

\begin{itemize}
    \item \textbf{Garment \& Wearable} (5 tasks): Virtual Try-On, Clothing Detail, Clothing Display, Universal Wearing, Pose Change.
    \item \textbf{Composition \& Layout} (3 tasks): Background Replace, Outpaint, Product Extract.
    \item \textbf{Text Operations} (3 tasks): Text Edit, Tagline Removal, Selling Point Display.
\end{itemize}

Table~\ref{tab:ecom-tasks} summarizes the definition and key properties of each task. The multi-reference tasks (Virtual Try-On and Universal Wearing) take two images as input — a model image and a product image — matching the paired-reference input format used during training.

\begin{table}[ht]
\centering
\caption{The 11 tasks of Ecom-Bench. $^\dagger$~denotes tasks with multi-image reference input (model image~+~product image). 100 test samples are used per task.}
\label{tab:ecom-tasks}
\small
\begin{tabular}{ll}
\toprule
\textbf{Task} & \textbf{Description} \\
\midrule
Virtual Try-On$^\dagger$         & Dress a model image with a specified garment               \\
Clothing Detail                   & Generate a zoomed-in detail view of garment texture        \\
Clothing Display                  & Clothe a virtual mannequin model with a target garment     \\
Universal Wearing$^\dagger$      & Apply any wearable product to a model image                \\
Pose Change                       & Alter the model's pose while preserving garment appearance \\
\midrule
Background Replace                & Swap the product backdrop with a specified scene           \\
Outpaint                          & Coherently extend the canvas of a product display image    \\
Product Extract                   & Segment and extract the product with clean edges           \\
\midrule
Text Edit                         & Modify or add text overlays on a product display image     \\
Tagline Removal         & Remove promotional taglines from product images       \\
Selling Point Display             & Compose a marketing poster with highlighted product features \\
\bottomrule
\end{tabular}
\end{table}

\subsubsection{Benchmark Curation}
\label{sec:ecom-construction}

Ecom-Bench is curated from diverse real and synthetic e-commerce imagery to cover varied products, presentation styles, and editing conditions. Candidate images are filtered for visual quality and task suitability, after which task-specific templates and VLM assistance are used to construct context-appropriate editing instructions. For the multi-reference tasks, Virtual Try-On and Universal Wearing, the benchmark pairs a model image with a target product image. All resulting image--instruction samples undergo a final quality audit before inclusion.

Each sample follows the format \texttt{\{edit\_image, prompt, info\}}, where \texttt{edit\_image} contains either one image or a list of images for multi-reference tasks. This consistent representation allows all 11 tasks to share the same inference and evaluation pipeline while retaining their task-specific input requirements.

\subsubsection{Evaluation Metrics}
\label{sec:ecom-metrics}

\paragraph{Evaluation Dimension Library.}
We define a library of 14 evaluation dimensions, consisting of 6 general dimensions applicable across all editing contexts and 8 e-commerce-specific dimensions that target domain-critical quality attributes.

\smallskip
\noindent\textbf{General dimensions (G):}
\begin{itemize}
    \item \textbf{G1 Instruction Compliance} — fidelity of the output to the editing instruction.
    \item \textbf{G2 Visual Naturalness \& Seamlessness} — naturalness of compositing boundaries and absence of visible blending seams.
    \item \textbf{G3 Physical \& Lighting Plausibility} — geometric and photometric coherence of shadows, reflections, and perspective.
    \item \textbf{G4 Non-edited Region Preservation} — pixel-level retention of regions outside the intended edit.
    \item \textbf{G5 Image Quality} — sharpness and freedom from noise and visual artifacts.
    \item \textbf{G6 Overall Aesthetics} — conformance to commercial presentation standards.
\end{itemize}

\smallskip
\noindent\textbf{E-commerce-specific dimensions (E):}
\begin{itemize}
    \item \textbf{E1 Product Identity Consistency} — preservation of the product's appearance and brand identity.
    \item \textbf{E2 Text Accuracy} — legibility, spelling correctness, and positioning of text elements.
    \item \textbf{E3 Texture \& Fabric Fidelity} — accuracy of reproduced material textures and patterns.
    \item \textbf{E4 Model Identity \& Pose Naturalness} — naturalness of the model's appearance and body posture.
    \item \textbf{E5 Garment Fit \& Silhouette} — realism of garment drape and fit on the body.
    \item \textbf{E6 Cutout Precision \& Edge Quality} — cleanliness and precision of segmentation boundaries.
    \item \textbf{E7 Key Selling Point Conveyance} — clarity with which the highlighted product features are emphasized.
    \item \textbf{E8 Product Presentation Completeness} — display of the complete product, free of unintended cropping or omission.
\end{itemize}

\paragraph{Per-task Dimension Assignment.}
Each task is assigned exactly four dimensions under a \textbf{2-general + 2-e-commerce-specific} configuration. Table~\ref{tab:ecom-dims} lists the dimension assignments for all 11 tasks.

\begin{table}[ht]
\centering
\caption{Evaluation dimension assignments for the 11 Ecom-Bench tasks.
The G and E prefixes denote general and e-commerce-specific dimensions, respectively.}
\label{tab:ecom-dims}
\small
\setlength{\tabcolsep}{4.5pt}
\begin{tabular}{lllll}
\toprule
\textbf{Task} & \textbf{D1} & \textbf{D2} & \textbf{D3} & \textbf{D4} \\
\midrule
Virtual Try-On         & E5 Fit\,\&\,Silhouette & E3 Texture    & G2 Seamless   & G3 Physical  \\
Clothing Detail        & E1 Product\,Id.         & E3 Texture    & G5 Quality    & G6 Aesthetics \\
Clothing Display & E3 Texture              & E4 Model      & G3 Physical   & G6 Aesthetics \\
Universal Wearing      & E1 Product\,Id.         & E4 Model      & G2 Seamless   & G3 Physical  \\
Pose Change            & E4 Model                & E5 Fit\,\&\,Silhouette & G1 Comply   & G3 Physical \\
\midrule
Background Replace     & G1 Comply               & E1 Product\,Id. & E8 Complete & G3 Physical  \\
Outpaint               & G4 Preserve             & E8 Complete   & E1 Product\,Id. & G2 Seamless \\
Product Extract        & E6 Cutout               & E1 Product\,Id. & G1 Comply   & G5 Quality   \\
\midrule
Text Edit              & E2 TextAcc.             & E1 Product\,Id. & G2 Seamless & G4 Preserve  \\
Tagline Removal      & G1 Comply               & E1 Product\,Id. & E8 Complete & G2 Seamless  \\
Selling Point Display  & E7 SellingPt.           & E2 TextAcc.   & G1 Comply     & G6 Aesthetics \\
\bottomrule
\end{tabular}
\end{table}

\paragraph{Scoring and Aggregation.}
Each dimension is rated on a 1--5 integer scale. Every score level is specified by a concrete behavioral rubric that describes what an output at that level looks like, rather than relying on abstract adjectives such as ``good'' or ``poor''; this behavior-anchored design reduces inter-run variance and score drift under a stochastic VLM judge. Evaluation is performed by Gemini 3.1 Pro Preview, which is presented with the reference image(s), the editing instruction, and the generated output.

The per-sample composite score is the geometric mean of the four dimension scores:
\begin{equation}
    S = \bigl(d_1 \cdot d_2 \cdot d_3 \cdot d_4\bigr)^{1/4}, \quad d_i \in \{1, 2, 3, 4, 5\}.
    \label{eq:geomean}
\end{equation}
We deliberately adopt the geometric mean over the arithmetic mean, as it penalizes dimension-level failures more severely. This behavior is consistent with commercial quality requirements, under which a single critical defect — garbled product text, loss of product identity, or an unnaturally fitted garment — renders an image unusable regardless of its scores on the other dimensions. Task-level scores are obtained by averaging $S$ over the samples of each task, and the overall Ecom-Bench score is the macro-average over the 11 tasks.

\paragraph{CTR Score.}
We use the CTR reward model described in \cref{sec:ctr-reward} to evaluate the predicted commercial attractiveness of generated images across all 11 Ecom-Bench tasks. Trained on product impression and click logs through CTR bucket classification followed by pairwise ranking refinement, the model provides a complementary measure of learned click preferences.

For comparison, we retain only test cases completed by all models included in the CTR evaluation and rank their outputs by the reward model's scores within each case. For each model, we report the total number of appearances at ranks 1, 2, and 3, with equal weight assigned to each rank. This statistic measures how frequently a model's output ranks among the top three under the learned CTR reward model; it does not represent measured CTR or actual click improvement.

\subsubsection{Evaluation Protocol}
\label{sec:ecom-protocol}

We use Gemini 3.1 Pro Preview as the automated evaluator under a task-specific prompting protocol. For each sample, the judge is presented with the reference image(s), the editing instruction, and the generated output, and assigns scores for the four dimensions selected for that task. Virtual Try-On and Universal Wearing provide both the model and product images as references. The dimension scores are combined and aggregated as described above to obtain the per-task and overall Ecom-Bench results.

\section{Evaluation}
\label{sec:evaluation}
\newcolumntype{K}{>{\centering\arraybackslash}X}

\begin{table*}[t]
\centering
\caption{Results on ImgEdit and GEdit under a common Gemini 3.1 Pro Preview judging protocol. All task columns and the overall score are reported. Within the open-source and closed-source groups separately, the best result is shown in bold and the second-best is underlined.}
\label{tab:imgedit-gedit-results}
\setlength{\tabcolsep}{1pt}
\tiny
\begin{tabularx}{\textwidth}{@{}l*{9}{K}|K@{}}
\multicolumn{11}{l}{\textbf{Panel A: ImgEdit (Gemini judge)}} \\
\toprule
\textbf{Model} & \textbf{Add} & \textbf{Adjust} & \textbf{Remove} & \textbf{Replace} & \textbf{Back.} & \textbf{Style} & \textbf{Extract} & \textbf{Action} & \textbf{Compose} & \textbf{Overall} \\
\midrule
Flux-2.0 & 4.21 & 3.61 & 4.07 & 4.36 & 3.78 & 4.56 & 3.13 & 3.23 & 2.26 & 3.69 \\
Joy-Edit & 4.34 & 4.06 & 4.25 & 4.25 & 4.00 & 4.94 & 4.40 & 3.10 & 3.44 & 4.09 \\
Joy-Edit-Plus & \underline{4.41} & \textbf{4.24} & 3.62 & 3.67 & 3.53 & 4.41 & 2.22 & \textbf{3.39} & 2.60 & 3.57 \\
LongCat & 4.28 & 4.07 & \underline{4.29} & 4.44 & \underline{4.05} & \textbf{4.98} & 4.38 & 3.00 & \underline{3.47} & 4.10 \\
FireRed-Edit & 4.36 & \underline{4.11} & \textbf{4.40} & \textbf{4.59} & 3.94 & \underline{4.96} & \textbf{4.84} & 2.89 & 2.90 & \underline{4.11} \\
Qwen-Edit-2511 & \textbf{4.44} & 3.63 & 3.86 & 4.20 & 3.47 & 4.82 & 4.27 & 2.73 & 3.01 & 3.83 \\
KwaiMind & 4.33 & 3.87 & 4.26 & \underline{4.47} & \textbf{4.24} & 4.89 & \underline{4.50} & \underline{3.27} & \textbf{3.52} & \textbf{4.15} \\
\midrule
Seedream & 4.53 & 4.26 & \underline{4.27} & \underline{4.52} & \underline{4.29} & 4.96 & 3.24 & \textbf{4.49} & \textbf{3.87} & \textbf{4.27} \\
NanoBanana-2 & \underline{4.54} & \underline{4.52} & 4.20 & 4.50 & 4.06 & \underline{4.98} & \underline{3.76} & 2.85 & \underline{3.38} & 4.09 \\
GPT-Image-2 & \textbf{4.68} & \textbf{4.60} & \textbf{4.34} & \textbf{4.74} & \textbf{4.52} & \textbf{5.00} & \textbf{4.37} & \underline{3.17} & 2.41 & \underline{4.20} \\
\bottomrule
\end{tabularx}
\vspace{3pt}
\begin{tabularx}{\textwidth}{@{}l*{11}{K}|K@{}}
\multicolumn{13}{l}{\textbf{Panel B: GEdit (Gemini judge)}} \\
\toprule
\textbf{Model} & \textbf{Back.} & \textbf{Color} & \textbf{Material} & \textbf{Motion} & \textbf{Human} & \textbf{Style} & \textbf{Add} & \textbf{Remove} & \textbf{Replace} & \textbf{Text} & \textbf{Tone} & \textbf{Overall} \\
\midrule
Flux-2.0 & 5.07 & 6.01 & 5.71 & 5.02 & 4.68 & 5.94 & 5.44 & 4.53 & 4.73 & 5.23 & \textbf{6.38} & 5.34 \\
Joy-Edit & 5.46 & 5.13 & 5.15 & 4.39 & 3.97 & 5.70 & 5.62 & 5.91 & 5.42 & \underline{6.26} & 5.47 & 5.32 \\
Joy-Edit-Plus & 3.25 & 4.29 & 3.65 & 3.72 & 3.38 & 4.12 & 4.58 & 2.73 & 3.14 & 3.95 & 4.56 & 3.76 \\
LongCat & 5.24 & 4.99 & 4.94 & 4.00 & 3.87 & 5.60 & 5.54 & \underline{6.09} & 4.95 & 6.17 & 5.35 & 5.16 \\
FireRed-Edit & \textbf{5.91} & 6.11 & 6.08 & 5.69 & \underline{5.92} & 6.17 & 5.83 & 6.07 & 5.35 & 5.85 & \underline{5.87} & 5.90 \\
Qwen-Edit-2511 & \underline{5.75} & \underline{6.15} & \underline{6.15} & \underline{6.18} & 5.83 & \underline{6.24} & \underline{6.20} & 6.00 & \underline{5.64} & 5.69 & 5.82 & \underline{5.97} \\
KwaiMind & 5.55 & \textbf{6.37} & \textbf{6.42} & \textbf{6.83} & \textbf{6.46} & \textbf{6.49} & \textbf{6.41} & \textbf{6.20} & \textbf{6.27} & \textbf{6.36} & 5.85 & \textbf{6.29} \\
\midrule
Seedream & 5.99 & 6.11 & 6.20 & 6.24 & 5.54 & 6.46 & 6.08 & 5.95 & 6.21 & 6.47 & 5.89 & 6.10 \\
NanoBanana-2 & \underline{6.38} & \underline{7.06} & \textbf{6.92} & \underline{7.19} & \underline{7.24} & \underline{6.97} & \underline{6.74} & \underline{6.85} & \underline{7.21} & \textbf{7.47} & \textbf{7.19} & \underline{7.02} \\
GPT-Image-2 & \textbf{6.83} & \textbf{7.19} & \underline{6.78} & \textbf{7.29} & \textbf{7.56} & \textbf{7.22} & \textbf{6.96} & \textbf{6.92} & \textbf{7.39} & \underline{7.25} & \underline{6.75} & \textbf{7.10} \\
\bottomrule
\end{tabularx}
\end{table*}

\begin{table*}[t]
\centering
\caption{Results on the English and Chinese splits of REDEdit. Within the open-source and closed-source groups separately, the best result is shown in bold and the second-best is underlined.}
\label{tab:rededit-results}
\setlength{\tabcolsep}{0.5pt}
\tiny
\begin{tabularx}{\textwidth}{@{}l*{15}{K}|K@{}}
\multicolumn{17}{l}{\textbf{Panel A: REDEdit English}} \\
\toprule
\textbf{Model} & \textbf{Add} & \textbf{Adjust} & \textbf{Back.} & \textbf{Beauty} & \textbf{Color} & \textbf{Compose} & \textbf{Extract} & \textbf{Portrait} & \textbf{Low} & \textbf{Motion} & \textbf{Remove} & \textbf{Replace} & \textbf{Style} & \textbf{Text} & \textbf{View} & \textbf{Overall} \\
\midrule
Flux-2.0 & 3.96 & 3.32 & 4.09 & \textbf{3.02} & 3.28 & 2.96 & 1.31 & \underline{3.79} & \textbf{3.66} & 4.30 & 3.43 & 4.12 & 4.41 & 2.75 & \textbf{2.95} & 3.42 \\
Joy-Edit & 3.95 & 3.32 & 3.90 & 2.29 & 3.58 & 3.01 & \underline{2.46} & 3.54 & \underline{3.11} & 3.86 & 3.64 & 4.10 & \underline{4.75} & 3.28 & 1.67 & 3.36 \\
Joy-Edit-Plus & 2.90 & 1.69 & 2.43 & 1.42 & 1.93 & 2.22 & 1.07 & 3.42 & 2.20 & 3.38 & 1.83 & 2.10 & 3.18 & 2.25 & 1.75 & 2.25 \\
LongCat & 4.02 & 3.25 & 3.91 & 2.31 & 3.55 & 2.97 & 2.32 & 3.49 & 2.98 & 3.91 & 3.62 & 4.20 & 4.69 & \underline{3.48} & 1.69 & 3.36 \\
FireRed-Edit & \textbf{4.37} & \textbf{3.75} & \textbf{4.27} & 2.82 & \textbf{3.92} & \underline{3.51} & 2.38 & 3.60 & 2.89 & 4.33 & \underline{4.10} & \textbf{4.33} & \textbf{4.77} & \textbf{3.66} & 2.39 & \underline{3.67} \\
Qwen-Edit-2511 & 4.20 & 3.21 & 3.69 & 2.76 & 3.19 & 3.22 & 2.26 & 3.77 & 2.52 & \textbf{4.61} & 3.63 & 4.10 & 4.57 & 3.22 & 2.57 & 3.43 \\
KwaiMind & \underline{4.27} & \underline{3.41} & \underline{4.10} & \underline{2.99} & \underline{3.85} & \textbf{3.61} & \textbf{3.13} & \textbf{3.99} & 3.04 & \underline{4.56} & \textbf{4.17} & \underline{4.27} & 4.68 & 3.45 & \underline{2.69} & \textbf{3.75} \\
\midrule
Seedream & 4.49 & 3.86 & \underline{4.35} & 3.48 & \textbf{4.29} & \underline{3.84} & 2.12 & \underline{4.47} & \underline{3.80} & 4.63 & \textbf{4.23} & \textbf{4.57} & \underline{4.91} & 4.30 & 2.55 & 3.99 \\
NanoBanana-2 & \underline{4.56} & \underline{3.91} & 4.11 & \textbf{4.07} & \underline{4.24} & 3.78 & \underline{2.79} & \textbf{4.60} & \textbf{4.24} & \underline{4.72} & 4.11 & 4.44 & 4.76 & \textbf{4.40} & \textbf{3.11} & \underline{4.12} \\
GPT-Image-2 & \textbf{4.73} & \textbf{4.03} & \textbf{4.45} & \underline{4.01} & 4.11 & \textbf{3.97} & \textbf{3.40} & 4.43 & 3.40 & \textbf{4.83} & \underline{4.19} & \underline{4.57} & \textbf{4.98} & \underline{4.40} & \underline{3.00} & \textbf{4.17} \\
\bottomrule
\end{tabularx}
\vspace{3pt}
\begin{tabularx}{\textwidth}{@{}l*{15}{K}|K@{}}
\multicolumn{17}{l}{\textbf{Panel B: REDEdit Chinese}} \\
\toprule
\textbf{Model} & \textbf{Add} & \textbf{Adjust} & \textbf{Back.} & \textbf{Beauty} & \textbf{Color} & \textbf{Compose} & \textbf{Extract} & \textbf{Portrait} & \textbf{Low} & \textbf{Motion} & \textbf{Remove} & \textbf{Replace} & \textbf{Style} & \textbf{Text} & \textbf{View} & \textbf{Overall} \\
\midrule
Flux-2.0 & 3.85 & 3.19 & 4.09 & \underline{2.79} & 3.25 & 2.92 & 1.26 & \textbf{4.06} & \textbf{3.66} & 4.15 & 3.06 & 3.88 & 4.38 & 2.64 & \underline{2.44} & 3.31 \\
Joy-Edit & 3.76 & 3.08 & 3.75 & 2.05 & 3.36 & 2.91 & 1.81 & 3.38 & 2.69 & 3.73 & 3.33 & 3.92 & 4.67 & 3.27 & 1.88 & 3.17 \\
Joy-Edit-Plus & 2.86 & 1.78 & 2.37 & 1.38 & 1.81 & 1.98 & 1.09 & 3.18 & 1.98 & 3.12 & 1.46 & 1.84 & 3.03 & 2.10 & 1.90 & 2.13 \\
LongCat & 3.72 & 3.09 & 3.71 & 2.01 & 3.45 & 2.78 & 1.73 & 3.46 & 2.71 & 3.79 & 3.31 & 3.88 & \underline{4.71} & 3.39 & 1.74 & 3.17 \\
FireRed-Edit & \textbf{4.33} & \textbf{3.62} & \textbf{4.17} & 2.62 & \textbf{4.05} & \underline{3.56} & \underline{2.43} & 3.64 & 2.96 & 4.46 & \underline{4.14} & \textbf{4.35} & \textbf{4.76} & \textbf{3.71} & 2.31 & \underline{3.67} \\
Qwen-Edit-2511 & 4.15 & 3.22 & 3.63 & 2.68 & 3.35 & 3.33 & 2.32 & 3.43 & 2.54 & \underline{4.47} & 3.75 & 4.15 & 4.61 & 3.40 & 2.38 & 3.43 \\
KwaiMind & \underline{4.26} & \underline{3.41} & \underline{4.13} & \textbf{2.84} & \underline{3.81} & \textbf{3.63} & \textbf{3.11} & \underline{3.79} & \underline{3.17} & \textbf{4.60} & \textbf{4.15} & \underline{4.34} & 4.62 & \underline{3.42} & \textbf{2.74} & \textbf{3.73} \\
\midrule
Seedream & 4.48 & \underline{3.78} & \underline{4.27} & 3.49 & 4.09 & \underline{3.82} & 1.74 & 4.45 & \underline{3.96} & 4.72 & 4.14 & \underline{4.60} & \underline{4.88} & 4.22 & 2.58 & 3.95 \\
NanoBanana-2 & \underline{4.59} & 3.72 & 4.17 & \underline{3.96} & \underline{4.22} & 3.80 & \underline{2.71} & \underline{4.48} & \textbf{4.39} & \underline{4.83} & \textbf{4.32} & 4.60 & 4.79 & \underline{4.24} & \underline{3.07} & \underline{4.13} \\
GPT-Image-2 & \textbf{4.67} & \textbf{3.93} & \textbf{4.34} & \textbf{4.02} & \textbf{4.32} & \textbf{4.08} & \textbf{3.65} & \textbf{4.67} & 3.41 & \textbf{4.86} & \underline{4.25} & \textbf{4.67} & \textbf{4.99} & \textbf{4.38} & \textbf{3.11} & \textbf{4.22} \\
\bottomrule
\end{tabularx}
\end{table*}

\begin{table*}[t]
\centering
\caption{Ecom-Bench visual scores and aggregate CTR ranking scores. VTO, Display, Wear, TLR, and Sell denote Virtual Try-On, Clothing Display, Universal Wearing, Tagline Removal, and Selling Point Display, respectively. For each of the 1,100 test cases, outputs are ranked by predicted CTR; rank-1, rank-2, and rank-3 appearances receive equal weight, and CTR reports their sum. Within the open-source and closed-source groups separately, the best result is shown in bold and the second-best is underlined. Missing results are denoted by ``--''.}
\label{tab:ecom-results}
\setlength{\tabcolsep}{0.7pt}
\tiny
\begin{tabularx}{\textwidth}{@{}l*{11}{K}|KK@{}}
\multicolumn{14}{l}{\textbf{Ecom-Bench visual score and CTR}} \\
\toprule
\textbf{Model} & \textbf{VTO} & \textbf{Display} & \textbf{Wear} & \textbf{Detail} & \textbf{Pose} & \textbf{Back.} & \textbf{Outpaint} & \textbf{Extract} & \textbf{Text} & \textbf{TLR} & \textbf{Sell} & \textbf{CTR} & \textbf{Overall} \\
\midrule
Flux-2.0 & 4.09 & \underline{3.75} & \underline{2.23} & \underline{3.04} & 3.79 & \underline{3.36} & \textbf{4.57} & 3.70 & 2.68 & \underline{3.46} & 3.65 & 293 & \underline{3.48} \\
Joy-Edit & -- & 3.61 & -- & 2.66 & 4.04 & 3.21 & 3.46 & \textbf{3.97} & \textbf{3.78} & 2.36 & \textbf{4.06} & -- & 3.46 \\
Joy-Edit-Plus & \underline{4.13} & 3.49 & 2.15 & 2.22 & 3.74 & 2.93 & 3.60 & 2.69 & 2.76 & 3.00 & 3.91 & \underline{406} & 3.15 \\
LongCat & -- & 2.63 & -- & 2.47 & 4.23 & 3.14 & 3.55 & 2.77 & 3.31 & 2.61 & 3.38 & -- & 3.12 \\
FireRed-Edit & \textbf{4.27} & 2.78 & 1.81 & 1.82 & \underline{4.33} & 3.27 & \underline{4.23} & 3.63 & 3.57 & 3.33 & 3.79 & 362 & 3.35 \\
Qwen-Edit-2511 & 3.62 & 3.21 & 1.72 & 2.62 & 4.14 & 2.87 & 3.92 & \underline{3.77} & 3.01 & 2.63 & 3.50 & 348 & 3.18 \\
KwaiMind & 3.98 & \textbf{4.19} & \textbf{2.39} & \textbf{4.25} & \textbf{4.44} & \textbf{3.56} & 2.55 & 3.69 & \underline{3.71} & \textbf{3.53} & \underline{4.06} & \textbf{544} & \textbf{3.67} \\
\midrule
Seedream & 4.13 & 4.31 & 2.43 & 3.54 & 4.27 & 3.19 & 2.63 & 3.02 & 3.88 & 2.60 & 4.15 & \textbf{525} & 3.47 \\
NanoBanana-2 & \underline{4.62} & \underline{4.41} & \underline{2.89} & \underline{3.97} & \underline{4.37} & \underline{3.64} & \textbf{4.11} & \underline{4.00} & \underline{4.34} & \underline{3.20} & \underline{4.54} & 391 & \underline{4.01} \\
GPT-Image-2 & \textbf{4.64} & \textbf{4.57} & \textbf{3.06} & \textbf{4.44} & \textbf{4.60} & \textbf{3.84} & \underline{3.53} & \textbf{4.28} & \textbf{4.38} & \textbf{4.10} & \textbf{4.75} & \underline{431} & \textbf{4.20} \\
\bottomrule
\end{tabularx}
\end{table*}

We compare KwaiMind with six open-source image editors: Flux.2-dev~\cite{flux-2-2025}, Joy-Image-Edit, Joy-Image-Edit-Plus~\cite{song2026joyai}, LongCat~\cite{longcat2025image}, FireRed-Image-Edit-1.0, and Qwen-Image-Edit-2511~\cite{wu2025qwen}, and three additional closed-source systems: Seedream5.0, NanoBanana-2, and GPT-Image-2. To make the comparison internally consistent, ImgEdit and GEdit are evaluated with the same Gemini 3.1 Pro Preview judge for every model, while REDEdit follows its original English and Chinese evaluation protocol. Ecom-Bench reports both the task-specific visual score defined in \cref{sec:ecom-metrics} and a CTR-based comparison over the test cases completed by all models included in the CTR evaluation. For each test case, the model outputs are ranked by their predicted CTR scores; appearances at rank 1, rank 2, and rank 3 are counted, and their equal-weight sum is reported for each model. In all tables, open-source and closed-source models are separated by a horizontal rule and ranked independently: the best score within each group is bolded and the second-best is underlined.

\subsection{General Image Editing Results}
\label{sec:eval-general}

Figure~\ref{fig:bench-general} gives an overview of the aggregate results, while Tables~\ref{tab:imgedit-gedit-results} and~\ref{tab:rededit-results} provide the complete task-level breakdown. KwaiMind achieves the best overall result among open-source models on all four general benchmarks: 4.15 on ImgEdit, 6.29 on GEdit, 3.75 on REDEdit English, and 3.73 on REDEdit Chinese. The improvements are not confined to a single edit family. On ImgEdit, KwaiMind leads the open-source group on background and compositional editing and ranks second on replacement and extraction. On GEdit, it obtains the strongest open-source result on material alteration, motion change, human editing, style change, subject removal, and subject replacement. This breadth indicates that continued pre-training preserves general editing competence despite the subsequent specialization toward e-commerce data.

The bilingual REDEdit results further show that the gain transfers across languages. KwaiMind ranks first among open-source systems on both English and Chinese overall scores, with particularly strong results for composition, extraction, portrait editing, motion, and removal. The English and Chinese scores are also close, suggesting that the bilingual instruction augmentation used during CT does not favor one language at the expense of the other. Closed-source systems remain stronger overall: GPT-Image-2 reaches 7.10 on GEdit and 4.17/4.22 on REDEdit EN/CN, while Seedream obtains the highest ImgEdit score of 4.27. The remaining gap is concentrated in several difficult categories, including viewpoint changes and some fine-grained appearance edits.

\subsection{Ecom-Bench Results}
\label{sec:eval-ecom}

Figure~\ref{fig:bench-ecom} summarizes the Ecom-Bench visual and CTR ranking results, while Table~\ref{tab:ecom-results} reports the per-task visual scores and the aggregate CTR ranking score. KwaiMind achieves an overall visual score of 3.67, outperforming every open-source baseline. KwaiMind leads the open-source group on Clothing Display, Universal Wearing, Clothing Detail, Pose Change, Background Replace, Tagline Removal, and the overall score. The largest margins occur on Clothing Detail and Clothing Display, where the model must preserve local appearance or transfer garments while maintaining identity and structure. These results align with the emphasis of the SFT data mixture and the fine-grained consistency reward.

Performance is less uniform on Outpaint and Product Extract. Flux.2-dev obtains the strongest open-source Outpaint score, while Joy-Image-Edit leads Product Extract. These categories indicate that specialization does not uniformly dominate strong task-specific priors and remain important directions for further improvement. Joy-Image-Edit and LongCat do not support the multi-image inputs required by Virtual Try-On and Universal Wearing and therefore lack results for those tasks; their reported overall values are macro-averages over nine available tasks rather than all eleven and should be interpreted with this limitation.

The closed-source systems retain a clear advantage in the visual evaluation. GPT-Image-2 achieves the strongest closed-source score on ten of the eleven tasks and an overall score of 4.20, followed by NanoBanana-2 at 4.01. KwaiMind nevertheless narrows the gap on several specialized tasks: its scores on Clothing Detail, Pose Change, and Clothing Display are close to the strongest proprietary results, demonstrating that targeted data and reward design can substantially improve production-oriented capabilities without relying on a closed model.

\paragraph{CTR comparison.}
Joy-Image-Edit and LongCat support only single-image inputs, so we exclude them from the CTR comparison. We evaluate the remaining eight models on all 1,100 test cases. For each case, the model outputs are sorted by their predicted CTR scores, and appearances at rank 1, rank 2, and rank 3 are counted separately. The three counts are equally weighted and summed to form the reported CTR ranking score. As shown in Figure~\ref{fig:bench-ecom}, KwaiMind records 215, 164, and 165 appearances at the three ranks, respectively, yielding the highest aggregate score of 544. This result complements the VLM-based visual score by measuring how consistently a model places among the most commercially attractive candidates under the learned preference model. Such a signal is valuable because a visually faithful edit is not necessarily an effective retail creative. Prior work on CTR-aware creative generation and product-poster optimization similarly uses behavioral feedback to guide generation beyond aesthetics alone~\citep{yang2024cg4ctr,fan2026autopp}; thus, KwaiMind's gain indicates stronger practical potential for producing deployable commercial imagery, rather than only higher judge-based visual quality.

\clearpage
\subsection{Visualization}
\label{sec:visualization}

\begin{figure}[H]
    \centering
    \includegraphics[width=\textwidth,height=0.78\textheight,keepaspectratio]{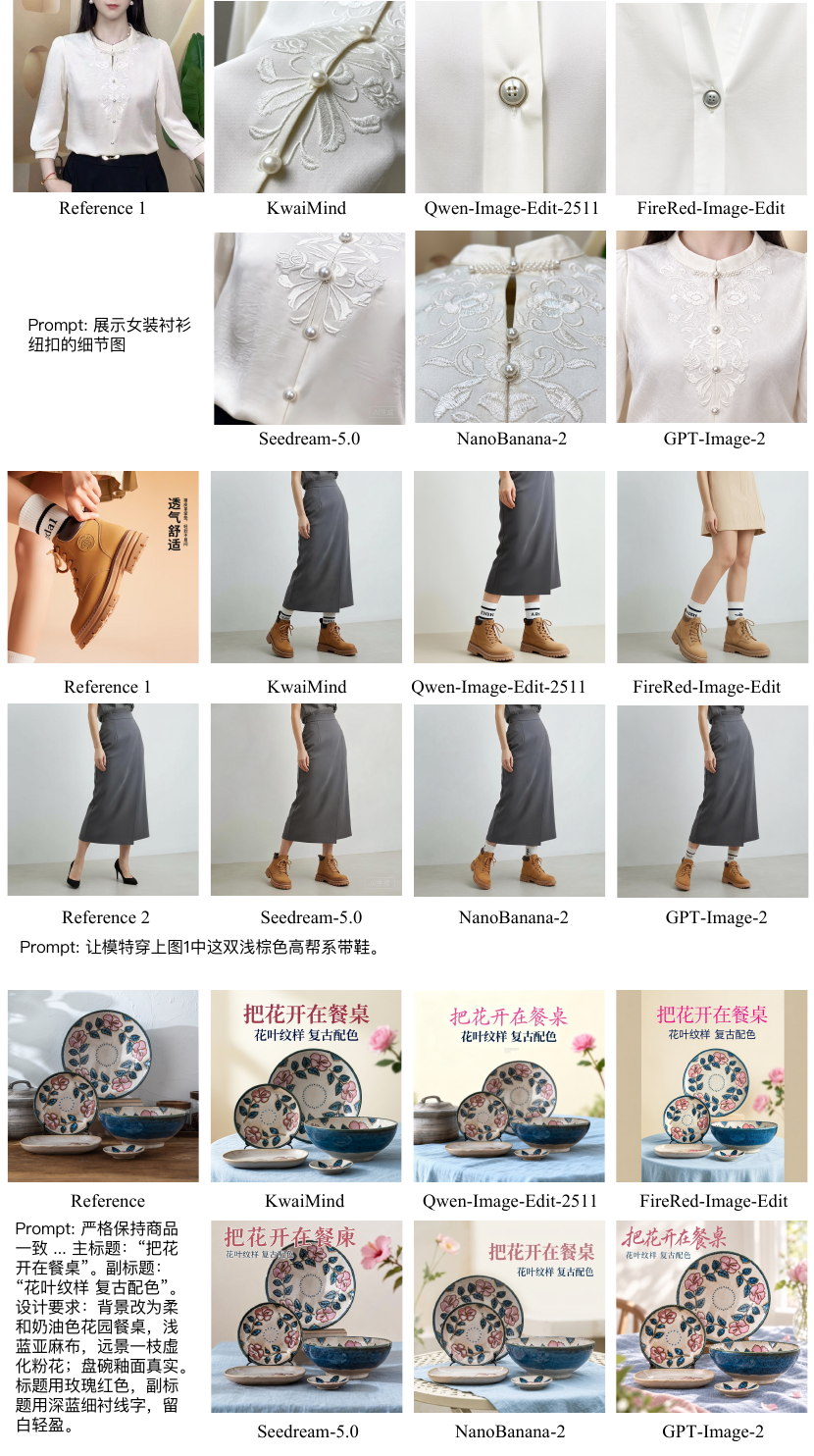}
    \caption{Qualitative comparisons on e-commerce image editing tasks (1/3).}
    \label{fig:ecom-visual-1}
\end{figure}

\begin{figure}[p]
    \centering
    \includegraphics[width=\textwidth,height=0.90\textheight,keepaspectratio]{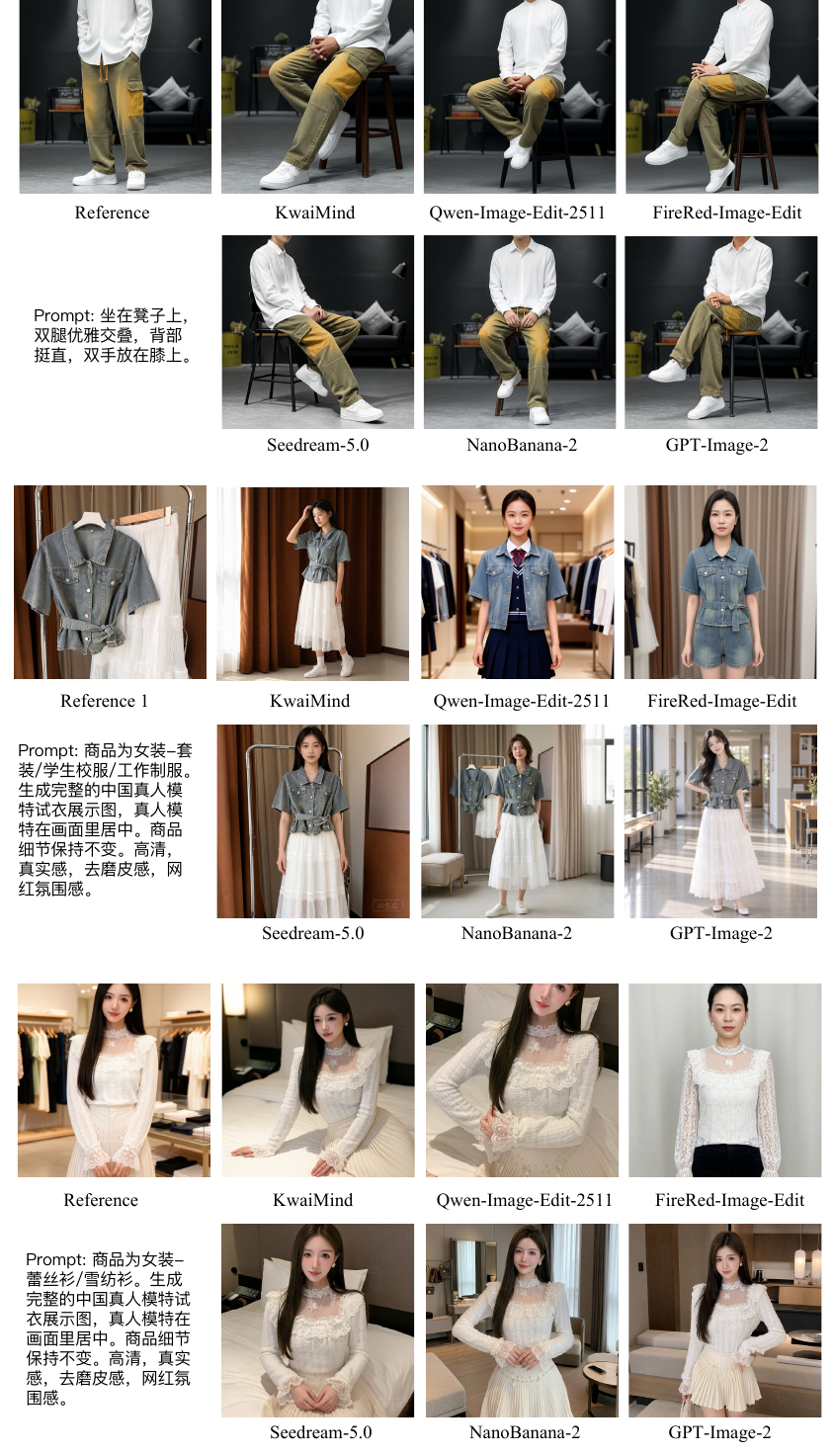}
    \caption{Qualitative comparisons on e-commerce image editing tasks (2/3).}
    \label{fig:ecom-visual-2}
\end{figure}

\begin{figure}[p]
    \centering
    \includegraphics[width=\textwidth,height=0.90\textheight,keepaspectratio]{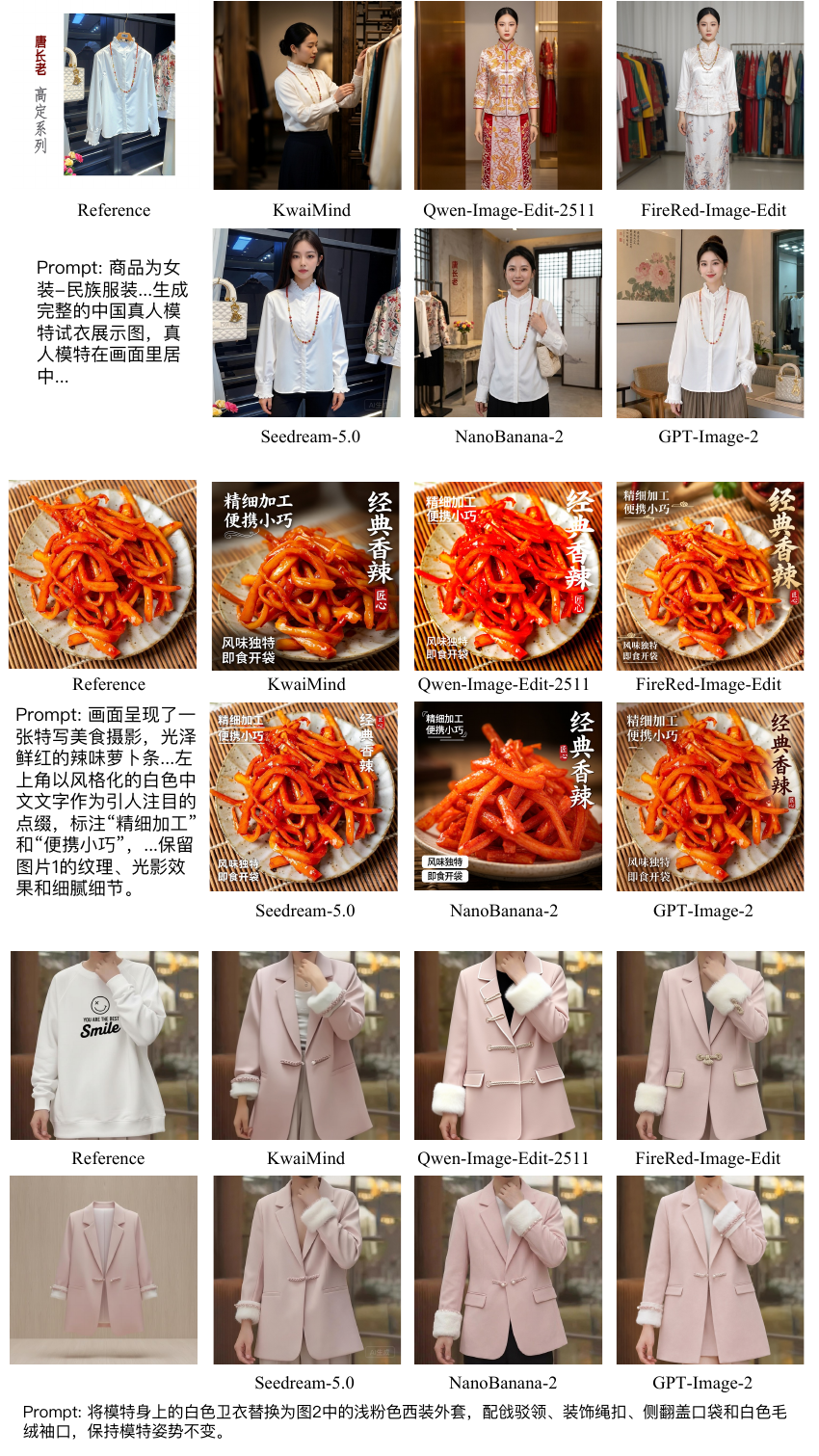}
    \caption{Qualitative comparisons on e-commerce image editing tasks (3/3).}
    \label{fig:ecom-visual-3}
\end{figure}

\begin{figure}[p]
    \centering
    \includegraphics[width=\textwidth,height=0.90\textheight,keepaspectratio]{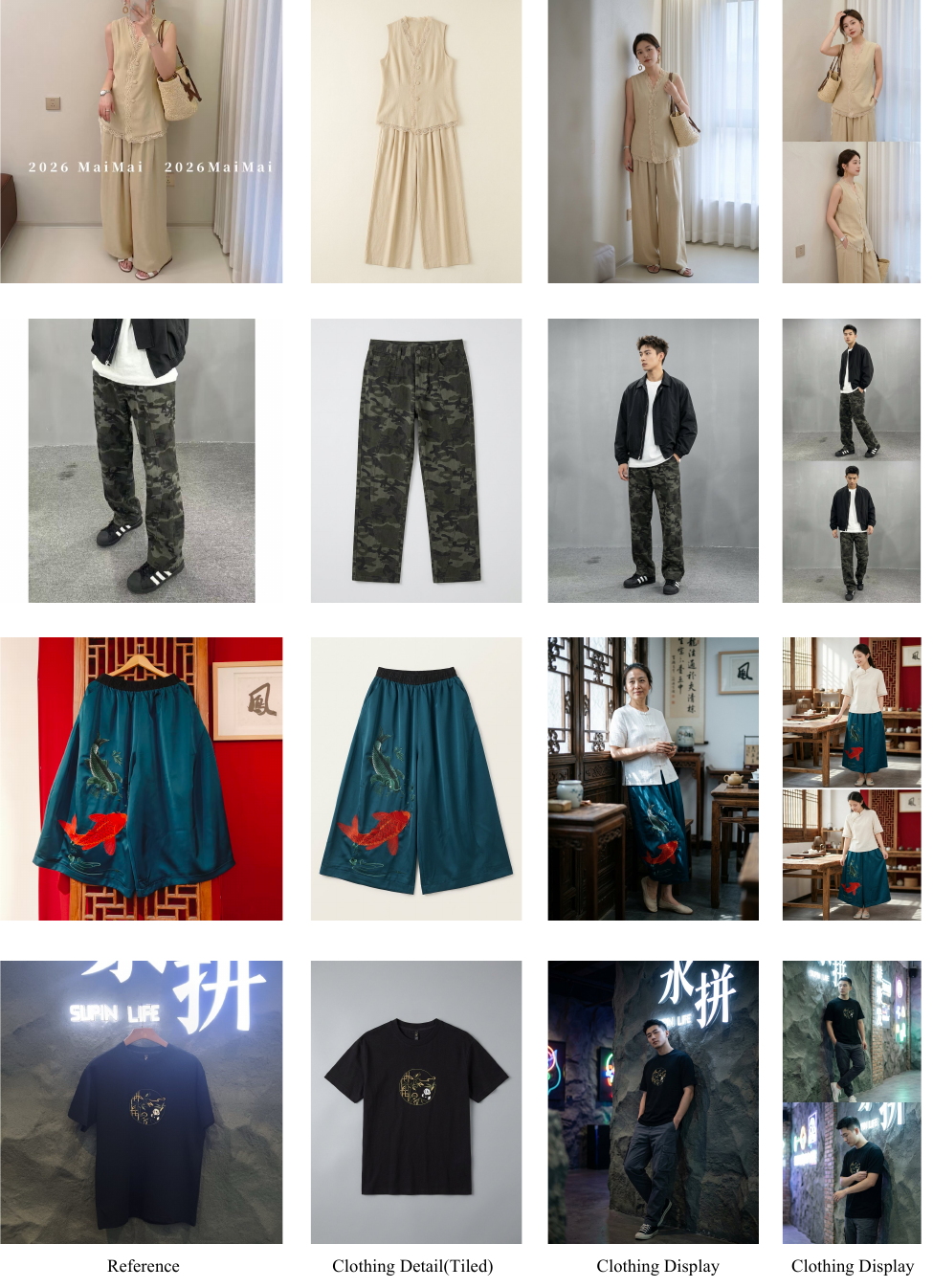}
    \caption{Garment presentation examples from KwaiMind, including tiled views and clothing displays on virtual models.}
    \label{fig:ecom-visual-4}
\end{figure}

\begin{figure}[p]
    \centering
    \includegraphics[width=\textwidth,height=0.90\textheight,keepaspectratio]{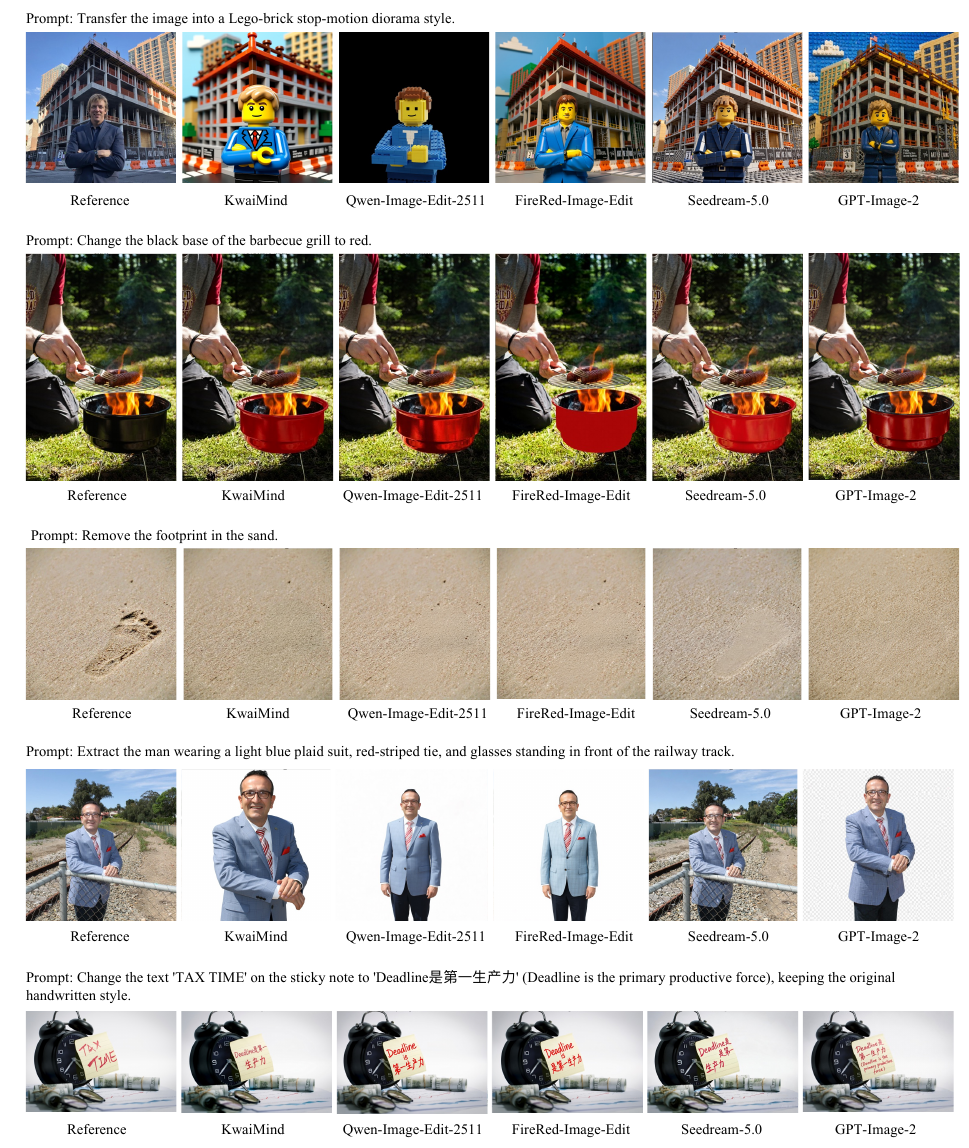}
    \caption{Qualitative comparisons on general image editing tasks (1/2).}
    \label{fig:general-visual-1}
\end{figure}

\begin{figure}[p]
    \centering
    \includegraphics[width=\textwidth,height=0.90\textheight,keepaspectratio]{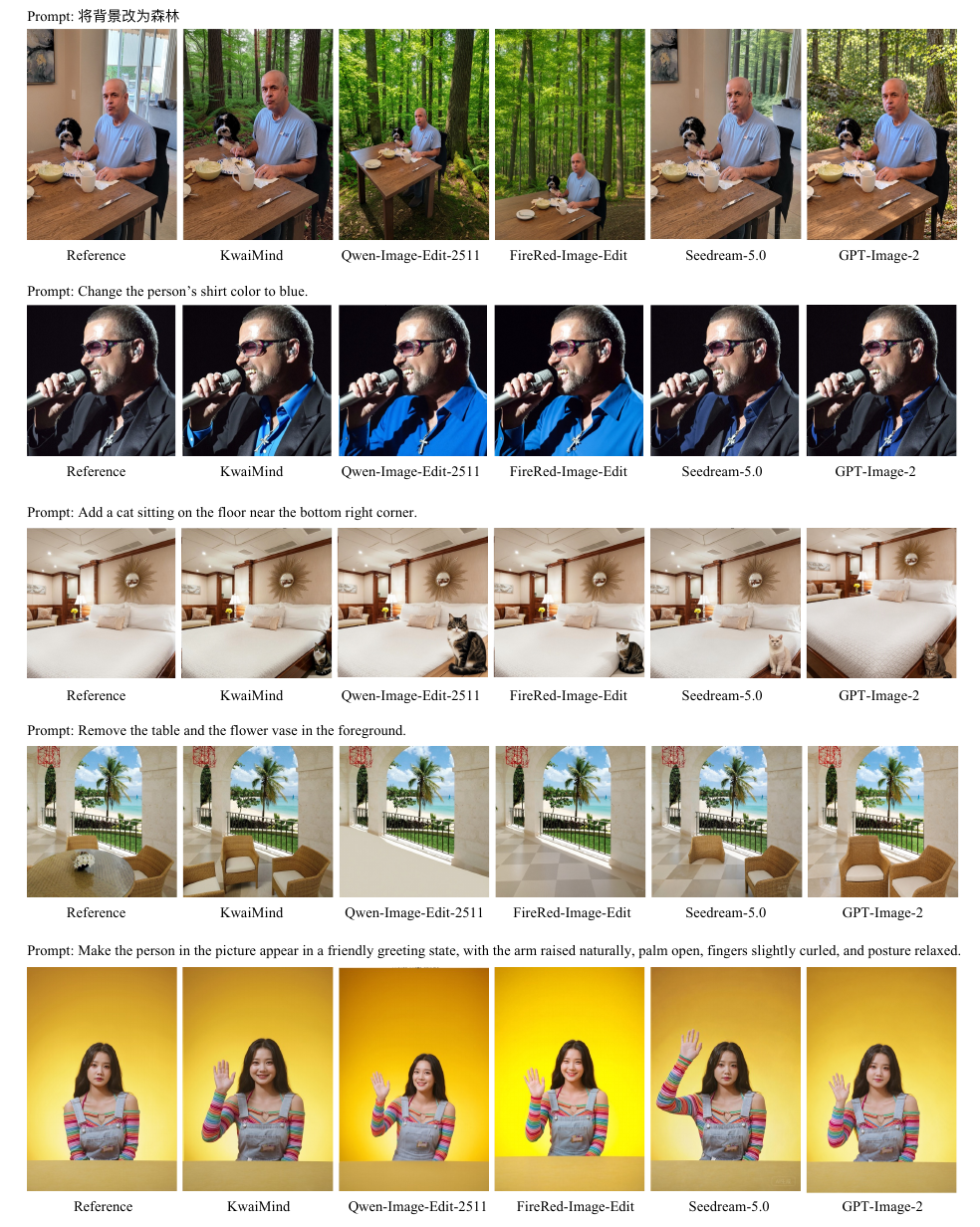}
    \caption{Qualitative comparisons on general image editing tasks (2/2).}
    \label{fig:general-visual-2}
\end{figure}

Figures~\ref{fig:ecom-visual-1}--\ref{fig:ecom-visual-3} and Figures~\ref{fig:general-visual-1}--\ref{fig:general-visual-2} present qualitative comparisons on e-commerce and general image editing tasks, respectively. Each comparison shows the reference image(s), editing instruction, and outputs from KwaiMind and representative baselines, complementing the quantitative results with a direct view of instruction following, content preservation, and visual quality across the two domains. Figure~\ref{fig:ecom-visual-4} provides additional garment presentation examples, including flat-lay views and clothing displays on virtual models.

\clearpage

\clearpage

\section{Conclusion}
\label{sec:conclusion}

We present KwaiMind, an image editing system that combines broad editing competence with the requirements of e-commerce content production through an agent-based data engine, staged training, and specialized reward optimization. Ecom-Bench complements general benchmarks with task-specific visual evaluation and CTR-based ranking. KwaiMind achieves the strongest aggregate results among the evaluated open-source editors across these benchmarks, while CTR-guided optimization and online material selection demonstrate practical commercial value. These findings highlight the benefit of aligning data, training objectives, and evaluation with real production needs. Future work will focus on closing the visual-quality gap with proprietary systems and improving robustness on compositional and multi-reference edits.

\section*{Contribution}
\label{sec:contributions}

\noindent\textbf{Core Contributors} (listed alphabetically): Boheng Zhang, Fan Yang, Jia Sun, Junlong Wu, Wenwu Ou, Yuting Hu, Zijun Li

\noindent\textbf{Major Contributors} (listed alphabetically): Dewen Fan, Fei Zuo, Honglie Wang, Huaiqing Wang, Pengcheng Wei, Yimin Zhou

\noindent\textbf{Support Contributors} (listed alphabetically): Haixuan Gao, Lihui Peng, Tingxuan She, Yuqing Li

\newpage

\bibliographystyle{abbrvnat}
\nobibliography*
\bibliography{bibtex}

\end{document}